\documentclass[review]{elsarticle}

\usepackage{graphicx}
\usepackage{amssymb}
\usepackage{times}
\usepackage{epsfig}
\usepackage{amsmath}
\usepackage{caption}
\usepackage{subcaption}
\usepackage{pifont}% http://ctan.org/pkg/pifont
\usepackage{url}
\usepackage{pifont}
\usepackage{multirow}
\usepackage[]{natbib}
\usepackage{algpseudocode}
\usepackage{color}
\usepackage{textcomp}
\usepackage{array}
\usepackage{algorithmicx}
\usepackage[ruled]{algorithm}
\usepackage[export]{adjustbox}
\usepackage{multirow}
\usepackage{graphicx}
\usepackage{hyperref}

\begin{document}

\begin{frontmatter}

%\title{Indic Word-level Script Identification using Offline-Online multi-modal Deep Network from Character-level Training Data}
\title{Indic Handwritten Script Identification using Offline-Online multi-modal Deep Network}

\author[label1,label7]{Ayan~Kumar~Bhunia} 
\author[label2]{Subham~Mukherjee} 
\author[label3]{Aneeshan Sain}
\author[label6]{Ankan~Kumar~Bhunia}
\author[label4]{Partha~Pratim~Roy}
\author[label5]{Umapada Pal}
\address[label1]{Institute for Media Innovation, Nanyang Technological University, Singapore.}
\address[label7]{Centre for Vision, Speech and Signal Processing, University of Surrey, England, United Kingdom.}
\address[label2]{Department of ECE, Institute of Engineering \& Management, Kolkata, India.}
\address[label3]{Department of EE, Institute of Engineering \& Management, Kolkata, India.}
\address[label6]{Department of Electrical Engineering, Jadavpur University.}
\address[label4]{Department of CSE, Indian Institute of Technology, Roorkee, India.}
\address[label5]{Computer Vision and Pattern Recognition Unit, Indian Statistical Institute, Kolkata, India.}

%\cortext[cor2]{Corresponding author: \href{mailto:2partharoy@gmail.com}{2partharoy@gmail.com}}

\begin{abstract} \label{introduction}
In this paper, we propose a novel approach of word-level Indic script identification using only character-level data in training stage. Our method uses a multi-modal deep network which takes both offline and online modality of the data as input in order to explore the information from both the modalities jointly for script identification task. We take handwritten data in either modality as input and the opposite modality is generated through intermodality conversion. Thereafter, we feed this offline-online modality pair to our network. Hence, along with the advantage of utilizing information from both the modalities, the proposed framework can work for both offline and online script identification which alleviates the need for designing two separate script identification modules for individual modality. We also propose a novel conditional multi-modal fusion scheme to combine the information from offline and online modality which takes into account the original modality of the data being fed to our network and thus it combines adaptively. An exhaustive experimental study has been done on a data set including English(Roman) and 6 other official Indic scripts. Our proposed scheme outperforms traditional classifiers along with handcrafted features and deep learning based methods. Experiment results show that using only character level training data can achieve competitive performance against traditional training using word level data.
 
\end{abstract}

\begin{keyword}
\texttt{Handwritten Script Identification, Deep Neural Network, multi-modal Learning, Offline and Online Handwriting, Character Level Training.} 
\end{keyword}

\end{frontmatter}

%\linenumbers

\section{Introduction}

Script identification \cite{Sahare2017,singh2015offline} is a key step of Optical Character Recognition (OCR) in multi-script documents. Script can be defined as a writing system consisting of a set of specific symbols and graphical shapes. Each script features specific attributes that distinguish it from other ones. Script identification is of utmost importance to understand handwritten documents automatically. Identification of handwritten script has gained prime importance in document image processing community; one of the reasons being global digitization of several handwritten scriptures and books. The basis of such script identification is the unique spatial relation among the strokes of a particular script, which helps in distinguishing the scripts from one another. 

\begin{figure}[!h!t!b]
	\begin{center}
		\fbox{\includegraphics[scale = 0.8]{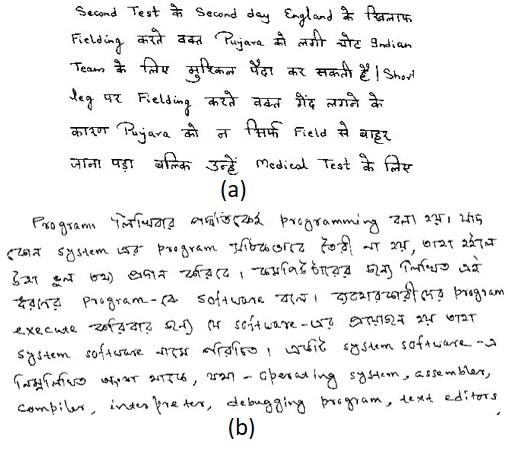}}
	\end{center}
	\centering
	\caption{Examples showing handwritten documents containing more than one script in a single document. (a) Devanagari and English (b) Bangla and English.}
	\label{fig:fig1}
\end{figure}

In multilingual \cite{singh2017handwritten} environment, script recognition gains significant importance, since every handwritten text recognition system is language specific. In a country like India, where there are 12 official scripts\footnote{\url{https://en.wikipedia.org/wiki/Languages_of_India}, accessed on 20/02/2018}, handwriting recognition becomes complicated. Hence, a robust script identification system is necessary to automate the process of text recognition \cite{ghosh2010script, ubul2017script}. The critical challenge encountered in such a task is that the handwritten text suffers from inherent challenges due to free flow nature of handwriting, unlike machine generated text which have fairly uniform structure. The variation in writing style among the individuals and complex shapes of characters are some of the major hindrances for handwritten script identification. Figure \ref{fig:fig1} shows some handwritten text documents written in different scripts. 

Most of the existing literature for script identification/ recognition focuses on extracting linguistic and statistical features at the word or the line level \cite{ghosh2010script,ubul2017script,chanda2009word,rajput2010handwritten}. On the contrary, the proposed framework is based on the hypothesis that the character set of a particular script alone contains distinctive features for the purpose of script recognition at word level. To extract the feature we employ both offline (image representation) as well as online (stroke order representation) modalities of handwritten text simultaneously, to utilize their combined potential for script identification task. To ensure the same, we have designed a deep neural architecture which can be trained in an end to end manner. To our knowledge, the proposed method is the first work which introduces the idea of combining offline and online modality in a single deep network to maximize the script identification task. Along with this multi-modal deep network, we also stress on the models trained using character-level data only and test the script identification task for both character and word level data.

\begin{figure}[!h!t!b]
	\begin{center}
		\fbox{\includegraphics[scale = 0.45]{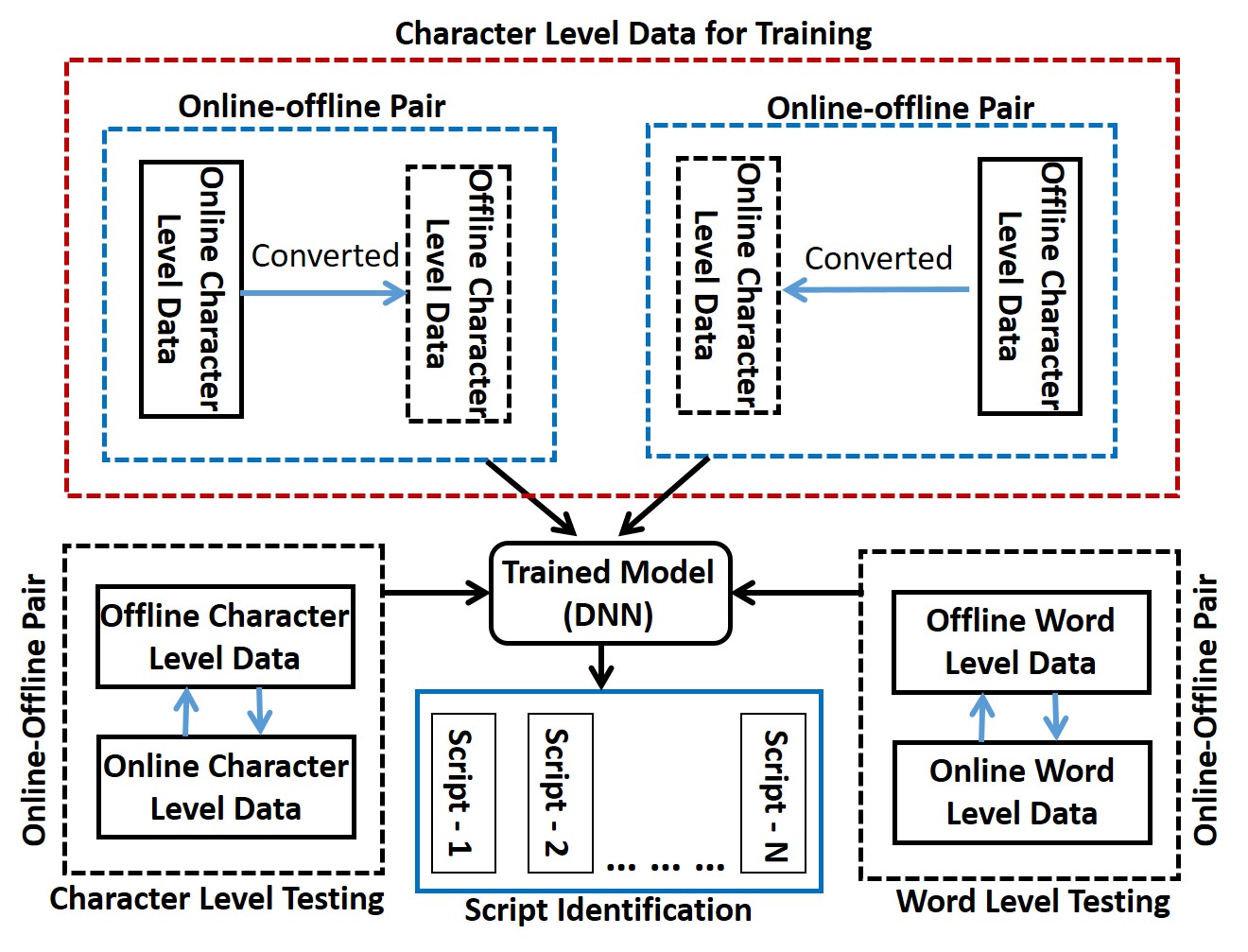}}
	\end{center}
	\centering
	\caption{Illustration of our proposed script identification framework. Character level data are used for training and testing is done for both character and word level data. We extract data in both online and offline modalities. If any of these two modality is missed, it is recovered from other one. }
	\label{fig:fig2}
\end{figure}

% Why Character Level Data
Only character level data in training stage provides us several advantages. Firstly, it is easier to collect large amount of character level data in lesser time, which saves both time and effort while preparing the dataset. Secondly, there are lesser number of unique characters compared to word level data, since several combinations of character sequence are possible for word level data. Thirdly, character level training is much lighter than that of word level data, thus making it ideal for employing in hand-held computing devices, e.g., mobile devices. It can be noted that character level data possesses a significant amount of information about the script which can be utilized to achieve word level script-identification performance. 

% Why Online Data is Important \\MERGED

The proposed deep network utilizes the information from both offline and online modalities jointly to leverage the respective advantages of both representations simultaneously. Online handwritten data \cite{namboodiri2004online} comprises of a 1-D signal in the form of sequence of points, whereas offline data comprises of images. The reason behind combining online modality is that it holds both spatial and temporal information of the character. On the contrary, the offline word images do not contain the temporal information. From the literature, it is noted that due to availability of dynamic information such as temporal order of the points, the performance of online system is comparatively better than the offline one \cite{hamdani2009combining}. However, offline data contains pixel level spatial information which partly complements the temporal and spatial information of online data.

Recently, the advent of convolutional deep architectures\citep{gomez2017improving} has made it possible to improve the performance of traditional feature-based approaches significantly. We draw our inspiration of the multi-modal framework from popular deep learning frameworks in RGB-D data \cite{xu2017multi,han2017cnns,wang2015mmss,asif2017multi} where it aims to utilize both RGB image and depth modality in a single deep network to get the advantage from both modalities simultaneously. Our proposed method is designed to handle both online and offline data in a single network rather than requiring different models. It eliminates the necessity of two different models for offline and online handwritten data individually. Our proposed framework first converts the online data into its offline form (and vice versa), thereafter feeding the data represented in two different modalities to the network, for learning the combined vector space for script identification.

In addition to this, we have proposed a novel conditional fusion scheme for script identification. The primary motivation is to include complementary information of offline and online modalities. For instance, if the original data is from online (offline) modality we convert it to its offline (online) equivalent, thereafter feeding both them into our deep network. During the multi-modal fusion, we combine features from both the modalities adaptively, by imposing the condition that the original data is from online modality. Note that, the data comes in single modality, i.e. both modalities of that particular data are not present. Thus, we perform inter modality conversion to generate the other one. Use of both modalities simultaneously provides us with two major advantages. Firstly, it provides better script identification performance; and secondly, this makes it possible to design a single system which can work for both offline as well as online handwritten data. The underlying idea here is that our proposed conditional multi-modal fusion mechanism would encourage the deep network to combine both the modalities considering their individual contributions.

%Main Contribution 
The main contributions of the paper are as follows. Firstly, we propose a deep multi-modal framework for script identification which uses online and offline modalities of character level data to exploit their shared information. We thus have a single framework which can handle both modalities by converting the data from the present modality to the other modality and learns a joint embedding to utilize the information from both the modalities for better performance. Here, we also develop a novel conditional multi-modal fusion scheme for effective combination of both the modalities. In addition to that, it avoids the need of requiring two different models to handle data in two different modalities. The trained model is used to identify the script for both online and offline character as well as word level data. To the best of our knowledge, this is the first attempt to develop a single framework which works in parallel for both online and offline data for handwritten script identification. Secondly, our proposed method enjoys the advantage of using light weight training model for script identification since we are using character level data for training which has fewer combinations compared to word level data. Thirdly, we have done an exhaustive experiments using a number of different scripts from different modalities, online and offline, to justify the feasibility and competitive performance with other existing baseline methods.

The rest of the paper is organized in the following manner. In Section 2 we discuss related work developed for script identification, stroke recovery task and popular deep multi-modal framework for different computer vision problems. In Section 3, the proposed framework for script identification has been described. In Section 4, we elaborate the experimental setup and discuss about the results of the various experiments conducted to justify the significance and efficiency of our proposed method. Finally, conclusions and future directions are given in Section 5.

\section{Related Work}

\textbf{Handwritten script identification:}  Handwritten script identification is an important task in developing a multilingual handwriting recognition system where more than one script might be present. A comprehensive survey on script identification has been presented in {\cite{ghosh2010script,ubul2017script}}. Various works have been reported on printed document script identification in \cite{chanda2009word}. However, script identification in handwritten documents is much more difficult compared to that in printed documents, due to varying handwriting styles of different individuals. A popularly explored approach in handwritten script identification is to extract linguistic and statistical feature followed by Support Vector Machine (SVM) as the classifier. In \cite{hiremath2010script}, such solution for word-level offline classification is proposed for Thai-Roman script classification. A technique for script identification in torn documents is proposed by Chanda et al. \cite{chanda2009word}, in which Roman and Indic scripts are considered to evaluate the performance. They worked with rotation invariant Zernike features and the rotation dependent gradient feature, using PCA-based methods to predict orientation and then apply an SVM classifier at the character level. The results are calculated for the word level using majority voting at the character level, followed by prediction at the document level in a similar fashion. Pal et al. \cite{pal2007handwritten} proposed a modified quadratic classifier using directional feature for recognition of off-line handwritten numerals of six popular Indic scripts. The bounding box of the numeric characters are divided into smaller blocks to capture the local information and directional feature is extracted at two levels, one from the original images and other one from the down sampled version of it through Gaussian Filtering. Moalla et al. proposed methods \cite{moalla2002extraction,moalla2004extraction} to separate out the Arabic text from the documents containing both Arabic and English words. In \cite{ferrer2014multiple}, Ferrer et al. has proposed a method for script identification in offline word images using word information index which estimates the amount of information included in a word. Different classifiers are trained using words with similar amount of information. During testing, the appropriate classifier is chosen based on the word information index of the query keyword. Regional local feature is studied in the work \cite{dhandra2007morphological}. In \cite{rajput2010handwritten},  Rajput et al. proposed a method based upon the features extracted using Discrete Cosine Transform(DCT) and Wavelets along with KNN classifier to identify seven  major scripts, namely, Devanagari, Gujarati, Gurumukhi, Kannada, Malayalam, Tamil, and Telugu at block level. Also, different textures feature have been explored for script identification task in various works \cite{hiremath2010script,hangarge2010offline,pal2012handwriting}.

Neural network based solutions are also popular for script identification as discussed in \cite{ghosh2010script}. In one of the earliest works, neural nets were employed for script identification in postal automation systems\cite{roy2005system,roy2005neural}. In \cite{sankaran2012recognition}, BLSTM is used for printed Devanagari Script recognition which uses five different features, namely, (a) the lower profile, (b) the upper profile, (c) the ink-background transitions, (d) the number of black pixels, and (e) the span of the foreground pixels. These features are fed to a Bi-RNN architecture using Connectionist Temporal Classification objective function which provides an improvement of more than 20\% Word Error Rate (WER) compared to the best available OCR system during its publication year.  In \cite{ul2015sequence}, a 1D-LSTM architecture, with one hidden layer is used for script identification at the text-line level to learn binary script models, and the reported identification accuracy for English-Greek scripts is 98.19\%.

Recently, Singh et al. \cite{singh2015word} proposed a word level script identification approach for handwritten images where they designed a set of 82 features using a combination of elliptical and polygonal approximation techniques. Authors considered a total of 7000 handwritten text words from six different Indic scripts - Bangla, Devanagarai, Gurumukhi, Malayalam, Oriya, Telugu and Roman script. They reported a maximum accuracy of 95.35\% using Multi-Layer Perceptron(MLP) classifier. Handwritten numeral script identification has been proposed by Obaidullah et al. \cite{obaidullah2015numeral}. Obaidullah et al. \cite{obaidullah2018handwritten} have done a comprehensive survey on handwritten indic script identification. Extreme Learning Machine(ELM) has been used for handwritten script identification in \cite{obaidullah2018extreme}. The work in \cite{obaidullah2019automatic} analyzed a particular question, which level of document (word, line, page) provides optimum performance for handwritten script identification. Authors concluded that line-level data provides most consistent performance among al, followed by page, block and word-level. Combination of different handcrafted feature extractors \cite{obaidullah2018automatic}, like directional stroke image component fractal dimension, structural and visual appearance, interpolation and Gabor energy based texture features, are used for line level indic-handwritten script identification.

Along with these, a few deep learning based approaches \cite{shi2016script,gomez2017improving,mei2016scene} for script identification in scene images have appeared in the literature recently. However, the potential of deep neural network for handwritten script identification has not been explored completely. In \cite{shi2016script}, local deep features are extracted using a pretrained CNN model and  discriminative clustering is carried out to obtain the mid level representation by learning a set of discriminative patterns from extracted local features. Following this, the deep features and the mid-level representations are jointly optimized in a deep network and their proposed model is termed as Discriminative Convolutional Neural Network(DisCNN). Gomez et al.\cite{gomez2017improving}  used the ensembles conjoined networks in order to learn from the stroke patches along with their relative importance.  

% \# offline to online Conversion REFINE

\textbf{Online Trajectory Retrieval:} Restoring of temporal order from offline handwriting has been worked since long \cite{doermann1995recovery,boccignone1993recovering}. Stroke and sub-stroke properties were utilized and authors provided a a taxonomy of local, regional and global temporal clues which were found to be beneficial for stroke recovery problem. In \cite{elbaati2009temporal}, Elbaati et al. proposed an approach to recover the stroke by segmenting the offline word image into strokes and labeling all the edges as successive parts of the strokes. Then, a Genetic Algorithm is applied to optimize these strokes and produce the best possible stroke order. An application of the above method is used in \cite{hamdani2009combining} which combines offline and online data for Hidden Markov Model(HMM) based Arabic handwriting recognition. The offline features and temporal stroke order from online data are complementary in nature, which in combination improve the recognition accuracy of the framework. In \cite{kato2000recovery}, Kato et al. proposed a stroke recovery technique which works for single stroke characters. This system labels each edge of the word image and connects them based on a predefined algorithm without any learning method. Very recently, Bhunia et al. \cite{kumarbhunia2018handwriting} proposed a deep-learning based solution using traditional sequence to sequence learning architecture for handwriting trajectory recovery. 

\textbf{Deep multi-modal learning:} multi-modal learning \cite{ngiam2011multi-modal, srivastava2012multi-modal} is a very popular concept in computer vision community in order to combine information from more than one sources. However, in document image analysis, more specifically for handwriting recognition task, there are hardly any application of multi-modal framework. In recent time, due to advancement of deep learning technology different modalities are combined for better accuracy in different problems like scene understanding, RGB-D object detection etc. In the task of Image Captioning \cite{karpathy2015deep} and Visual Question Answering \cite{ilievski2017multi-modal}, the language feature and image feature are combined using deep neural network architecture. In RGB-D data, both image and depth modalities are explored for various tasks like Object Recognition\cite{wang2015mmss}, Scene classification\cite{zhu2016discriminative}, Object Detection \cite{xu2017multi}. Given a vast literature for multi-modal learning, we attempt to use a deep multi-modal framework to explore the joint information from both offline and online handwritten data for script identification task.

\section{Proposed Framework}

\begin{figure}[!h!t!b]
	\begin{center}
		\fbox{\includegraphics[scale = 0.5]{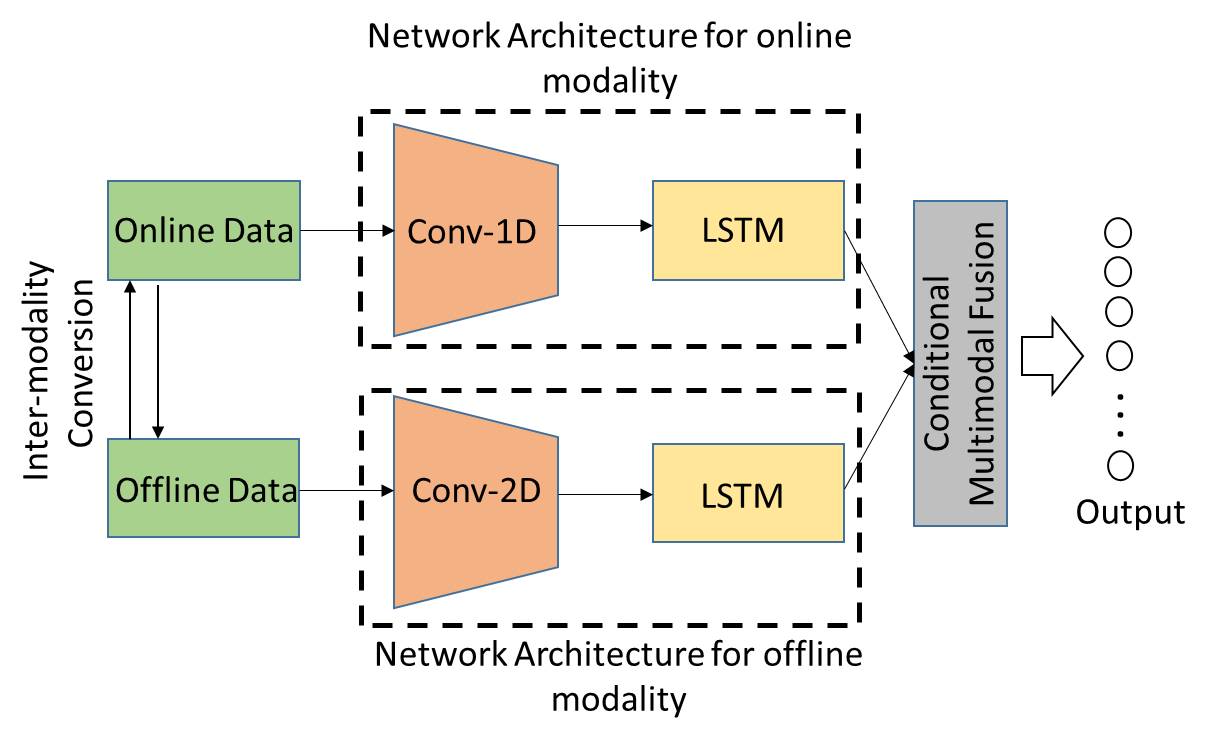}}
	\end{center}
	\centering
	\caption{The outline of our proposed multi-modal architecture for script identification. There are two inputs in our network, one from offline modality and other from online modality. The absent modality will be generated through intermodality conversion.}
	\label{fig:fig3}
\end{figure}

The proposed approach can be divided into two steps. In the first step, we extract the data from original modality to its equivalent opposite modality. In the next step, the data from both the modalities are considered simultaneously as input to deep neural network to combine their information. Note that only character level data is used to train the network. During testing, it can identify the script for both character and word level data from both modalities. In our work, we design Convolutional-LSTM architecture where Convolutional Network intends to extract more robust sequential feature from the data and LSTM module captures the contextual information of the sequence for better performance. An overview of our proposed multi-modal framework is given in Figure \ref{fig:fig3}. In this section, we describe our different modules of our framework serially.

\subsection{Inter-Modality Conversion}

Inter modality conversion is a key step in our framework. Since our proposed deep network takes the offline-online pair as the input, a modality-conversion is required in order to get the opposite modality from the input data.

\subsubsection{Offline to Online Conversion:}

\begin{figure}[!h!t!b]
	\begin{center}
		\fbox{\includegraphics[scale = 0.40]{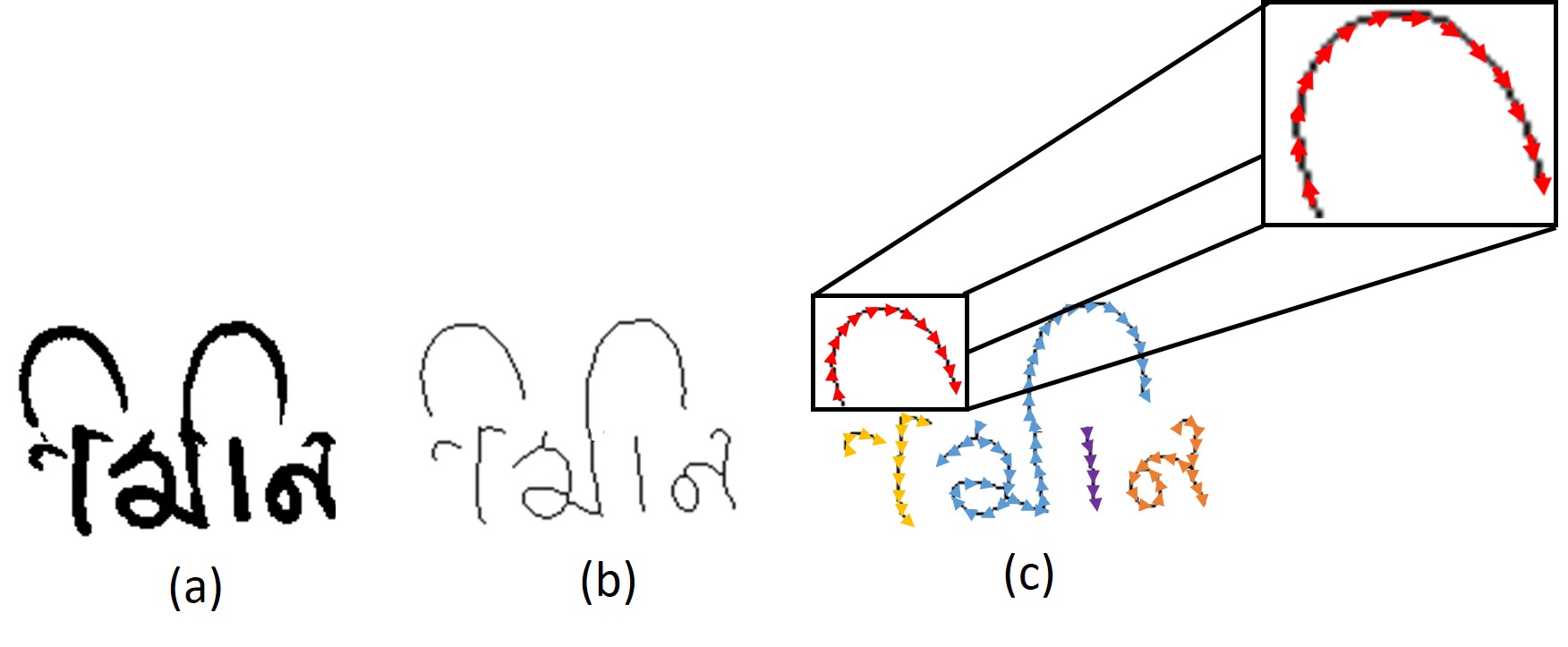}}
	\end{center}
	\centering
	\caption{Example showing stroke recovery for offline to online conversion. (a) Input offline image (b) Skeleton of that image (c) Direction of recovered stroke is shown using arrow. Different color signifies independently detected strokes.  }
	\label{fig:fig9}
\end{figure}

Offline to online conversion of handwritten data has been performed using \citep{nel2005estimating}. First, the skeleton image of the handwritten text is extracted through a thinning process to extract a parametric curve. In order to perform a comparison between a static image skeleton and a dynamic exemplar, translation is performed until the centroids of the two become aligned and following this, it is converted into a static image. The authors have thickened the static image skeleton and also the image obtained from the dynamic exemplar to a line width of approximately five pixels. After this, matching between the two images is carried out followed by trajectory extraction. The method that has been used for determining the pen trajectory from a static and a normalized image by making use of a Hidden Markov Model. For the problem of stroke order recovery, the sequence of states of the HMM are used to describe the sequence of pen positions as the image is produced. The HMM model is built from the skeleton of the static image. The dynamic exemplar is matched to the static image by making use of the Hidden Markov Model. An example of offline to online conversion is shown in Figure \ref{fig:fig9}.

\subsubsection{Online to Offline Conversion:} 
\begin{figure}[!h!t!b]
	\begin{center}
		\fbox{\includegraphics[scale = 0.35]{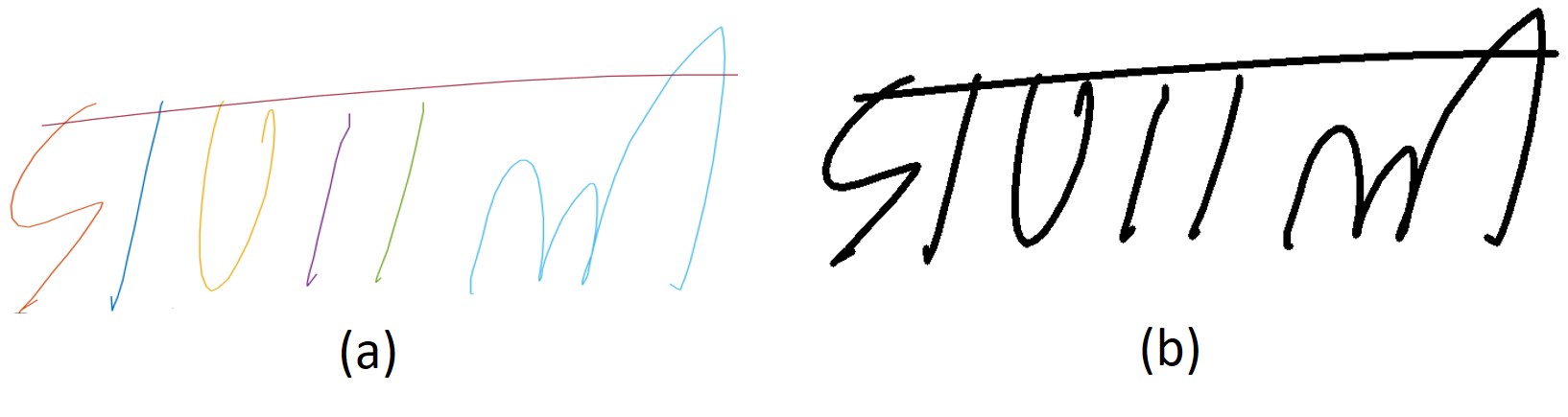}}
	\end{center}
	\centering
	\caption{Example showing conversion of online to offline modality. (a) Redrawn skeleton from the online coordinate points. (b) Obtained Offline representation after morphological thickening}
	\label{fig:fig8}
\end{figure}

Conversion of online data to offline is a trivial one. Online data consists of consecutive co-ordinate points representing the flow of writing. To convert the online data to its offline equivalent, we first define an empty image matrix based on the difference of maximum and minimum $(x,y)$ coordinate values. Thereafter, we mark those pixel positions of the empty image matrix based on the online coordinate points and join  consecutive points serially. This process generates the skeleton image of the handwritten data. Following this, a morphological thickening operation is performed in order to make the equivalent offline word image similar to the real offline word image data. Online to offline conversion is shown in Figure \ref{fig:fig8}

\subsection{Network Architecture for Online modality}

Online modality of handwritten data consists of sequential $(x,y)$ co-ordinate points representing the flow of writing. One of the naive ways could be to feed these sequential $(x,y)$ co-ordinate points to a LSTM module and get the state of the last time step as the final feature representation of the online data. However, in this proposed approach, we have designed a Convolutional LSTM architecture for the online stream of our network. The key idea of using CNN is to achieve a certain extent of shift, scale and distortion in-variance. For a 2D image, the local connectivity of CNN learns the correlation among neighbouring pixels. The objective of using CNN on the sequential co-ordinate points before feeding it to the LSTM module is to learn the correlation among the neighbouring co-ordinate points. In addition to that, it intends to achieve a certain degree of distortion or shift invariance which may arise due to free flow of writing of different individuals, and during capturing of co-ordinate points by sensors. Hence, convolutional network will convert the $(x,y)$ co-ordinate points into a high dimensional space by incorporating the spatial correlation among nearby sequential points in order to make it less variant to different individuals' handwriting and distortions. 

For every online data sample, there are N sequential points represented by two values, $ X $ co-ordinate value and $ Y $- co-ordinate value for each point. However, in order to formulate the 1-D convolution over these points, we consider these points as 1-dimensional signal with 2 channels, i.e. $D \in \mathbb{R}^{N\times 1 \times 2}$. We employ a 1-D convolution with filter $W \in \mathbb{R}^{M\times Q \times 1\times P}$ ($M$ filters with dimension of $ {Q\times 1\times P}$). We restrict the length of the filter $Q$ to 5. $P$ is the number of channels or feature maps in the input. For example, $P = 2$ for the first convolution layer since the input is 1-dimensional 2 channel having a length N. The output $G \in \mathbb{R}^{N\times 1 \times M}$ of 1D-convolution is given by

\begin{equation}
{G = Conv_{1D}(D;W)}
\centering
\end{equation} 

\begin{figure}[!h!t!b]
	\begin{center}
		\fbox{\includegraphics[scale = 0.30]{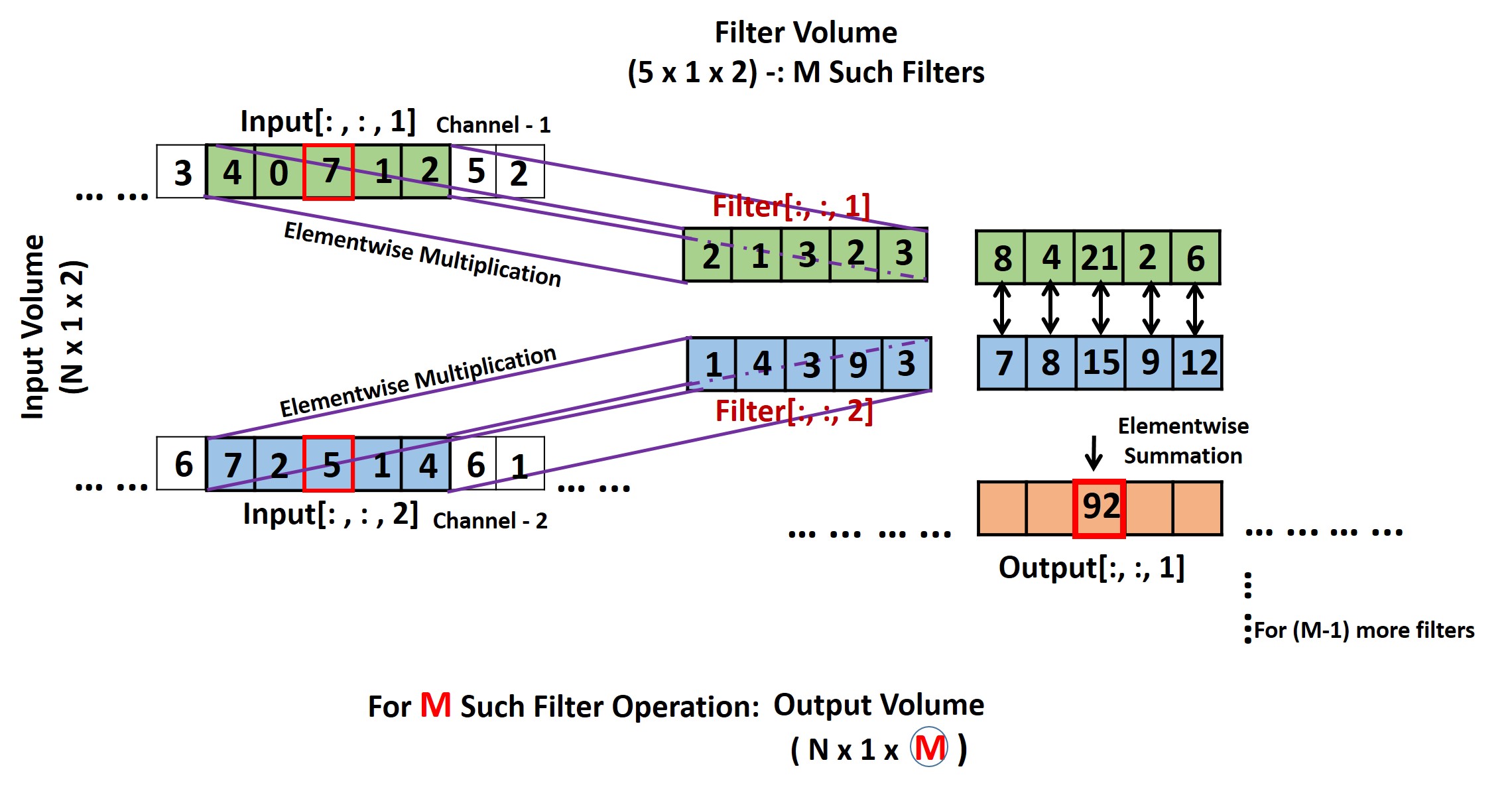}}
	\end{center}
	\centering
	\caption{Graphical illustration of 1D-convolution on online coordinate points}
	\label{fig:fig7}
\end{figure}

1-D convolution operation follows the same rule as that of 2D convolution with a small difference, which is that the filter used here is of size $L \times 1$ and it strides over only one direction(here time direction). A graphical illustration of 1D-convolution over the online coordinate points is shown in Figure \ref{fig:fig7}. In order to introduce the invariance against free flow writing of different individuals and noisy acquiring of data from sensors, we have added one maxpooling operation along the time direction after the second convolution layers. The window size of maxpooling operation is $2\times 1$, hence it reduces the number of data points to half.

However, note that, online data  represents the flow of writing with time. Hence, it is observed that more than one maxpooling operation reduces the performance. Inspired from the network of \cite{engelmann2017exploring}, we have used global maxpooling \cite{lin2013network} operation to obtain the global feature vector which contains the holistic information of all the co-ordinate points. This global feature vector is concatenated with the feature map of every data point and passes through to last convolution layer. This convolution layer intends to combine each point wise feature with a global feature adaptively. In our online stream, we have used six convolution layers with the number of filters being 32, 64, 128, 256, 256, and 512 respectively. Hence, we obtain an output tensor of size  $\mathbb{R}^{(N/2) \times 1 \times 512}$, which thereafter is to be fed to the LSTM module. Following this, we create a custom `Map-to-Sequence' layer as the bridge between convolutional layers and recurrent layers as mentioned in \cite{shi2016end}. This `Map-to-Sequence' layer converts the 3-dimensional tensor to a 2-dimensional time distributed feature representation of size $\mathbb{R}^{(N/2) \times 512}$ for the online data points. 

\subsection{Network Architecture for offline modality}

The online data was already time distributed, containing successive co-ordinate points which represent the flow of writing. In contrast to this, the offline data does not have any time information. Hence, in order to feed this to a LSTM module, we need to convert the offline word image into sequential feature representation. In handwriting recognition, one of the popular approaches is to use sliding based \cite{roy2016hmm, BHUNIA201812} feature extraction. However, the handcrafted features has its own limitation. To solve this problem, we have used a convolutional neural network\cite{shi2016end} in order to extract the sequential feature from offline images which thereafter is fed to the LSTM module. There may exist several cases where character from one script may resemble the character of a different script. For those instances, it will be beneficial to look at the contexts of those ambiguous characters. Hence, we employ a Convolutional-LSTM architecture\cite{shi2016end} for the script identification task where convolutional network is used to convert the offline image into its sequential deep convolutional feature representation. Then the LSTM module is used to capture the contextual information with in a sequence. It can take input images of arbitrary width which is one of the major requirements of our framework since we are training the network using character level data and predict the result for both character and word level data as well. Word level data usually has much longer width and there is large variation in the length depending on the number of characters present in that word. Resizing the width to a fixed size is not a good choice since it distorts the word image and it may eliminate some good script specific information. However, it is needed to scale all the images to a fixed height to feed them in the network keeping the aspect ratio same. 

\begin{figure}[!h!t!b]
	\begin{center}
		\fbox{\includegraphics[scale = 0.40]{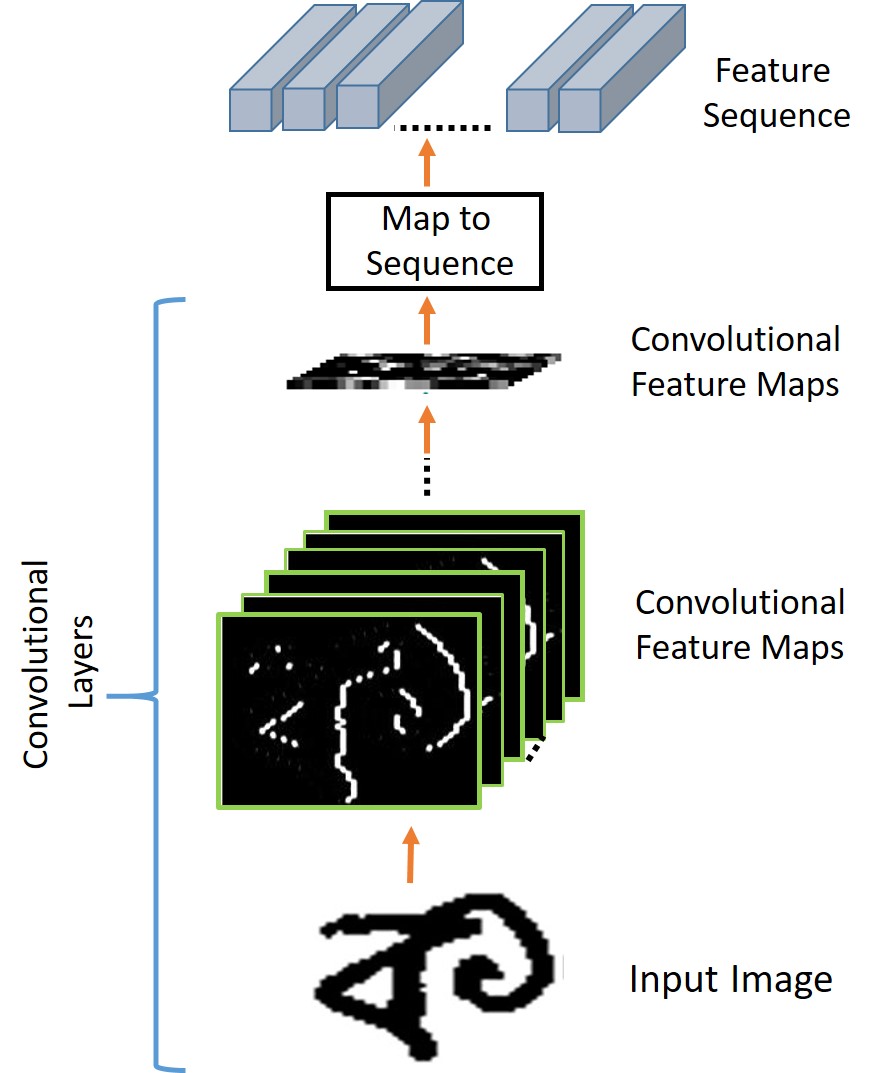}}
	\end{center}
	\centering
	\caption{Image to Feature Sequence Conversion}
	\label{fig:fig6}
\end{figure}

Generation of each feature vector of a feature sequence is done in a left-to-right manner on the feature maps, taken column wise. This denotes that the the concatenation of the i-th columns of all the maps gives rise to  the i-th feature vector. As per the architecture, the width of each such column is maintained at one pixel. Features are translation invariant due to the fact that layers of convolution, max pooling and element wise activation function operate in local neighborhood. Hence, each column of the feature maps actually maps to a specific area of the original image. Such regions are found in the same sequence to their corresponding columns on the feature maps from left to right. Each vector in the feature sequence can thus be regarded as a local image descriptor. Figure \ref{fig:fig6} graphically shows the process of feature sequence generation using convolutional architecture \cite{shi2016end}. Our convolutional architecture for feature sequence extraction is composed of seven convolutional layer. The major change we have done in our network is to include a global maxpooling \cite{lin2013network} operation to get a global feature vector for the sequence which is concatenated with every left-to-right feature map and are fed to one last convolutional layer before converting it to final feature sequence using `Map to Sequence' operation. The primary objective of including the global feature in our method is that besides gathering the local information, we can also consider the holistic representation of the entire image.

\begin{figure}[!h!t!b]
	\begin{center}
		\fbox{\includegraphics[scale = 0.25]{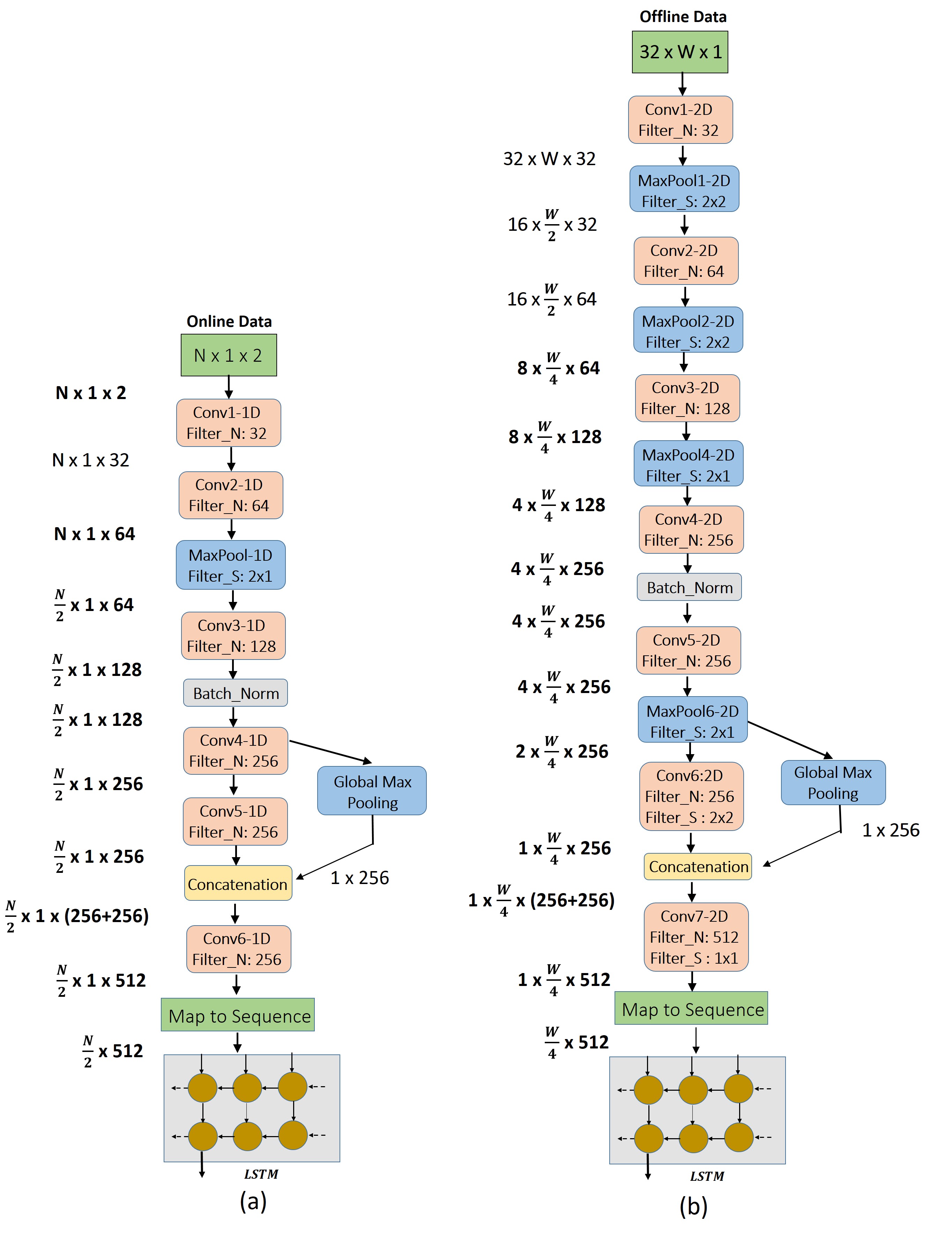}}
	\end{center}
	\centering
	\caption{The architecture for (a) online stream and (b) offline stream. The window size(filter size) for both pooling and convolution is 5x1 and 2x2 in case of online and offline stream respectively until mentioned specifically in the diagram. \texttt{Filter\_S} and \texttt{Filter\_N} stand for number of filters and size of the filter respectively.}
	\label{fig:fig4}
\end{figure}

\subsection{Conditional Multi-modal Fusion}

After obtaining the time distributed feature sequence from both offline and online stream of every data sample using 1-D and 2-D Convolutional Network, we feed those sequential features to two different LSTM (Long Short Term Memory)modules for each modality. Traditional RNN suffers from the problem of vanishing and exploding gradient. To overcome these drawbacks, a different type of RNN is used known as LSTM (Long Short Term Memory). A memory cell along with three multiplicative gates constitute an LSTM. These gates are called input gate, forget gate and output gate. From the conceptual point of view, the past contents are stored in memory cells while the input and output cells are used to enable the cell to store contents for a long period of time. The forget gate is used to clear the memory in the cell. The main advantage of an LSTM is its ability to handle better long term dependency. The core idea of using LSTM module is to extract the feature from cell state of last time step after LSTM has seen the complete sequence of the offline or online feature sequence. This leads to the consideration of sequential relation among all feature vectors of a sequence. This is highly expected in a task like script identification where two different scripts might have a few related characters which possess a certain extent of similarity. However, using the sequential approach, we can avoid this confusion by considering global representation including the sequential relation among successive feature vectors of a sequence.
For combination of offline and online information, we proposed a multi-modal conditional fusion method. Simple concatenation of the features of the modalities results redundant information decreasing the performance of the model. Also, it is necessary to take the most relevant information from the two modalities for correctly classifying the script. Thus, we used a novel fusion technique that dynamically assigns weights across the modalities representations. It learns the correlations between offline and online modality along with their adaptive contribution in fusion method. 

Let the feature from the final time step of the LSTM network be $F_{online}$ and $F_{offline}$ with size $\mathbb{R}^{1\times K}$ for online and offline modality respectively. At first, we concatenate $F_{online}$ and $F_{offline}$ to obtain $F_{concat}$ of size $\mathbb{R}^{1\times 2K}$. 

\begin{equation}
{F_{concat}^{1\times2K} = Concat(F_{online}^{1\times K}, F_{offline}^{1\times K})}
\centering
\end{equation} 

The concatenated feature representation is conditioned on a 2 bit binary vector $Z$ representing the original modality of the input data. It is important to let the model know the actual form of the input data, whether it is online data or offline data. It allows the model to give the priority to the original modality adaptively in calculating the final feature representation. After feeding $Z$ to the concatenated representation we get a feature vector $F_{cond}$ of size $\mathbb{R}^{(1\times 2K)+2}$. Thereafter, we pass it through a fully connected layer($FC$) of weights $W \in \mathbb{R}^{(2K+2) \times K}$. The primary objective of such fully connected layer is to learn the correlation between the two modalities in order to assign their respective weightage accordingly. The output of this fully connected layer is $F_{fc}$ with size $\mathbb{R}^{1\times K}$. Finally, the sigmoid function outputs the weight parameters. Using equation \ref{eq_Poff} and \ref{eq_Pon} we get the weights $P_{Offline}^{1\times K}$ and $P_{Offline}^{1\times K}$. These weights are element-wise multiplied with their corresponding modality representations. The final feature vector is obtained by adding these two feature representation. Then, a fully connected layer is used which has the same number of neuron as the number of classes. A softmax layer outputs the probability distribution of the script over the classes. The conditional fusion is carried out by the following operations.

\begin{equation}
{F_{fc}^{1\times K} = FC(F_{cond}^{1\times 2K+2} ; W_{fc,1}^{2K+2 \times K}) }
\centering
\end{equation} 

\begin{equation} \label{eq_Poff}
{ P_{Offline}^{1\times K} =  Sigmoid(F_{fc}^{1\times K}) }
\centering
\end{equation} 

\begin{equation} \label{eq_Pon}
{P_{Online}^{1\times K} = 1 - P_{Offline}^{1\times K} }
\centering
\end{equation} 

\begin{equation}
{{F_{offline,weighted}^{1\times K}} = F_{offline}^{1\times K}  \odot P_{offline}^{1\times K}}
\centering
\end{equation} 

\begin{equation}
{{F_{online,weighted}^{1\times K}} = F_{online}^{1\times K}  \odot P_{online}^{1\times K}}
\centering
\end{equation} 

\begin{equation}
{ {F_{Fusion, Final}^{1\times K}} = {F_{offline,weighted}^{1\times K}} + {F_{online,weighted}^{1\times K}} }
\centering
\end{equation} 

\begin{figure}[!h!t!b]
	\begin{center}
		\fbox{\includegraphics[scale = 0.28]{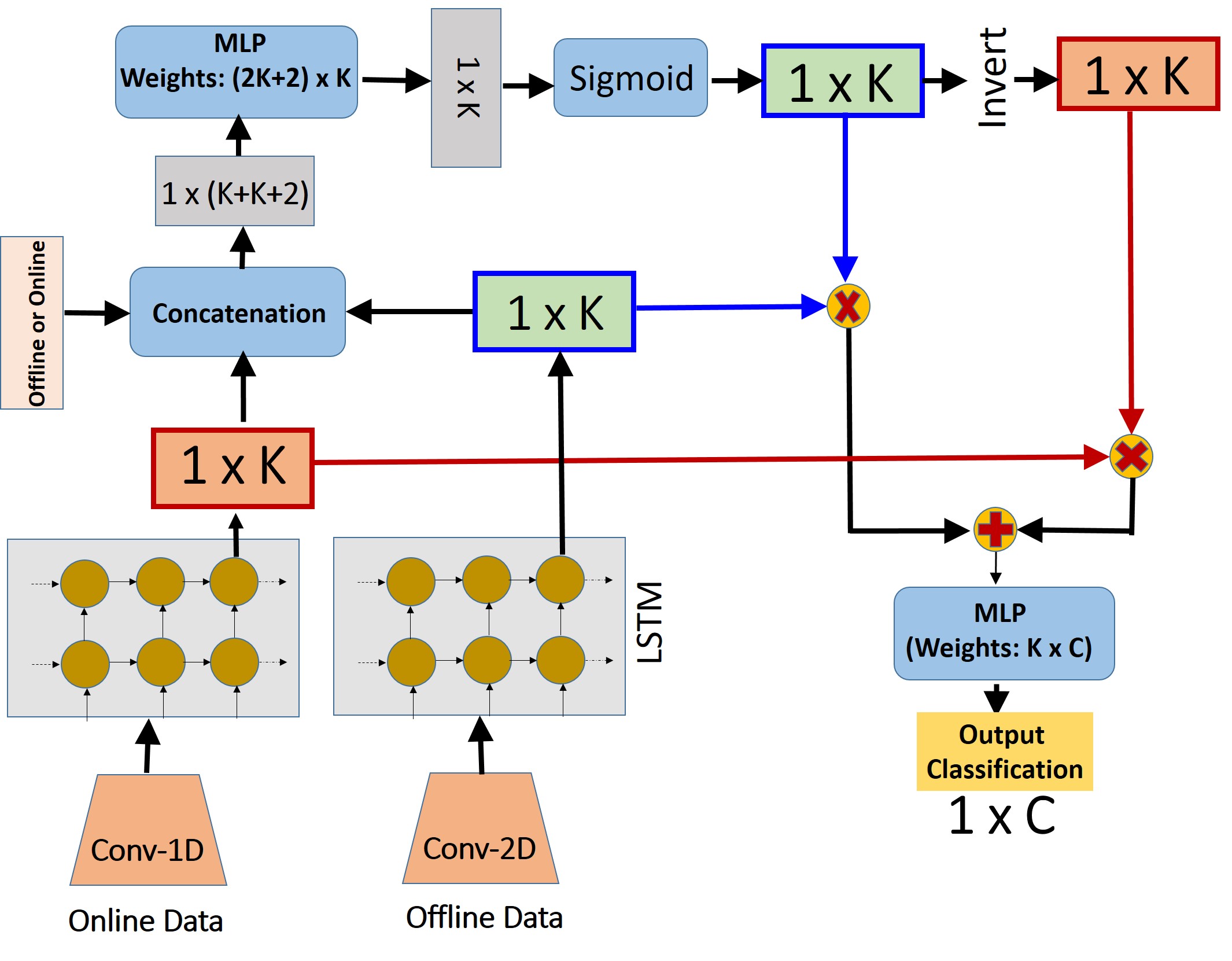}}
	\end{center}
	\centering
	\caption{Conditional multi-modal fusion scheme}
	\label{fig:fig5}
\end{figure}

\subsection{Implementation Details} \label{impl}

Our proposed multi-modal network uses character level data during training and it predicts the script identification result for both character and word level data. Let $(X_{i},Y_{i})$ be the online-offline modality pair for a particular sample data which is to be fed as the input to our network. It is to be noted that only one of the modalities, i.e. either $X_{i}$ or $Y_{i}$  is present originally, and we convert the other one using the method mentioned in section 3.1. $(L_{i}, Z_{i})$ are the two given labels to train the network in a supervised manner. Here, $L_{i}$ is the corresponding script label and $Z_{i}$ is the original modality from which the original data sample was fetched. $Z_{i}$  is the extra supervision we use to impose a condition during multi-modal fusion. During training, $L_{i}$ is used to calculate the cross-entropy loss for classification to train the network through back-propagation. During testing, the network predicts the identified script $L_{i}$ as output. However, $Z_{i}$ is present during both training and testing for conditional fusion of two different modalities. 

The network architecture of our framework is shown pictorially in Figure \ref{fig:fig4}. We have included one global average pooling operation in both offline and online stream networks in order to capture the holistic information about the data sample. The architecture for offline stream consists of 7 convolutional layers and 4 maxpooling layers. For the 3rd and 4th maxpooling layers, the filter size was fixed at 2x1, in order to get the feature maps with larger width, thus generating a larger feature sequence, which was found to be beneficial for capturing the spatial dependency among characters of words. We normalize every offline data to a height of 32 keeping the aspect ratio same. In order to feed the character level offline data in batches during training, we resize every character to a size of 32x32. However, we can feed offline images of arbitrary width to keep the aspect ratio constant during testing, but only one at a time.  This is expected because word images(during testing) usually have much longer width, i.e. much higher aspect ratio compared to single characters. On the contrary, for online data it has no such font size limitation, since it is already time distributed. The architecture of online stream consists of 6 convolutional layer and one maxpooling operation. In order to accelerate the training process, we have added two batch normalization layers in both of our offline and online stream networks. Next, we have used two layers LSTM network with 512 hidden LSTM units. Hence, K equals to 512 in our framework. The final feature for offline and online modality is extracted from the cell state of the last time step of two LSTM modules, after which multi-modal conditional fusion is carried out. 

We have implemented our complete system using Python and Tensorflow framework in 2.50 GHz Intel(R) Xeon(R) CPU, 32GB RAM and an NVIDIA Titan-X GPU. The weights of the model are initialized according to the Xavier initializer. All convolution and fully connected layers use Rectified Linear Units (ReLU). The training is carried out using Stochastic Gradient Descent algorithm with a momentum of 0.9 and learning rate 0.01. The network converges after 30K iterations with a batch size of 32. The learning rate is multiplied by 0.1 when the validation error stops decreasing for enough number of iterations. The weight decay regularization parameter is set to $5\times 10^{4}$.

\section{Experiments}

In this section, we report the performance of our script identification framework. We first introduce the datasets used for our study and then present the detailed script identification performance along with different baseline methods, error analysis and discussions.

\subsection{Datasets}
As per our findings, there exists no such standard datasets for handwritten script recognition. Here, we have collected various publicly available word and character recognition datasets \cite{BHUNIA201812,roy2016hmm} of different scripts to prepare our required database for script recognition. In our experiments, we have considered a total of 7 scripts for the performance evaluation, namely, Devanagari, Bangla, Odia, Gurumukhi, Tamil, Telugu and English. Among these scripts, Devanagari, Bangla, Gurumukhi and Odia are descended \cite{ghosh2010script} from the common ancestor script in the Brahmi script family. There exist a good extent of similarity between Bangla and Devanagari, Devanagari and Gurumukhi as mentioned in \cite{BHUNIA201812}. Similarly, Tamil and Telugu are two south Indian scripts. On the other side, English is a global language which is a medium of communication for different parts of the world. Most of the documents are bi-script which contains one of the regional languages with English. Hence, our selection of the scripts for performance evaluation is based on the intention to make the task of script identification more difficult. Table \ref{tab1} gives the detail of our dataset used for script recognition. All the experiments for script recognition have been done in a 10 fold cross validation mode with 7:2:1 training, validation and testing.  By this, 70\% data of dataset was used for training, 10\% data for validation and 20\% data for testing.

% \usepackage{multirow}
\begin{table}[]
\centering
\caption{Details of our script identification dataset}
\label{tab1}
\begin{tabular}{|c|c|c|c|c|}
\hline
\multirow{3}{*}{\textbf{Scripts}} & \multicolumn{4}{c|}{\textbf{Number of Samples}}                        \\ \cline{2-5} 
                                  & \multicolumn{2}{c|}{Character Level} & \multicolumn{2}{c|}{Word Level} \\ \cline{2-5} 
                                  & Online           & Offline           & Online         & Offline        \\ \hline
Bangla                            & 10,509             & 10,897              & 10,589           & 10,687           \\ \hline
Devanagari                        & 10,897             & 11,021              & 10,847           & 10,645           \\ \hline
Gurumukhi                         & 9,897             & 9,789              & 9,569           & 9,657           \\ \hline
Odia                              & 9,457             & 9,789              & 9,147           & 9,234           \\ \hline
Tamil                             & 9,456             & 9,476              & 9,874           & 9,476           \\ \hline
Telugu                            & 10,486             & 10,789              & 10,687           & 10,694           \\ \hline
English                           & 10,489             & 10,879              & 11,023           & 11,458           \\ \hline
\end{tabular}
\end{table}
  
\begin{figure}[!h!t!b]
	\begin{center}
		\fbox{\includegraphics[scale = 0.35]{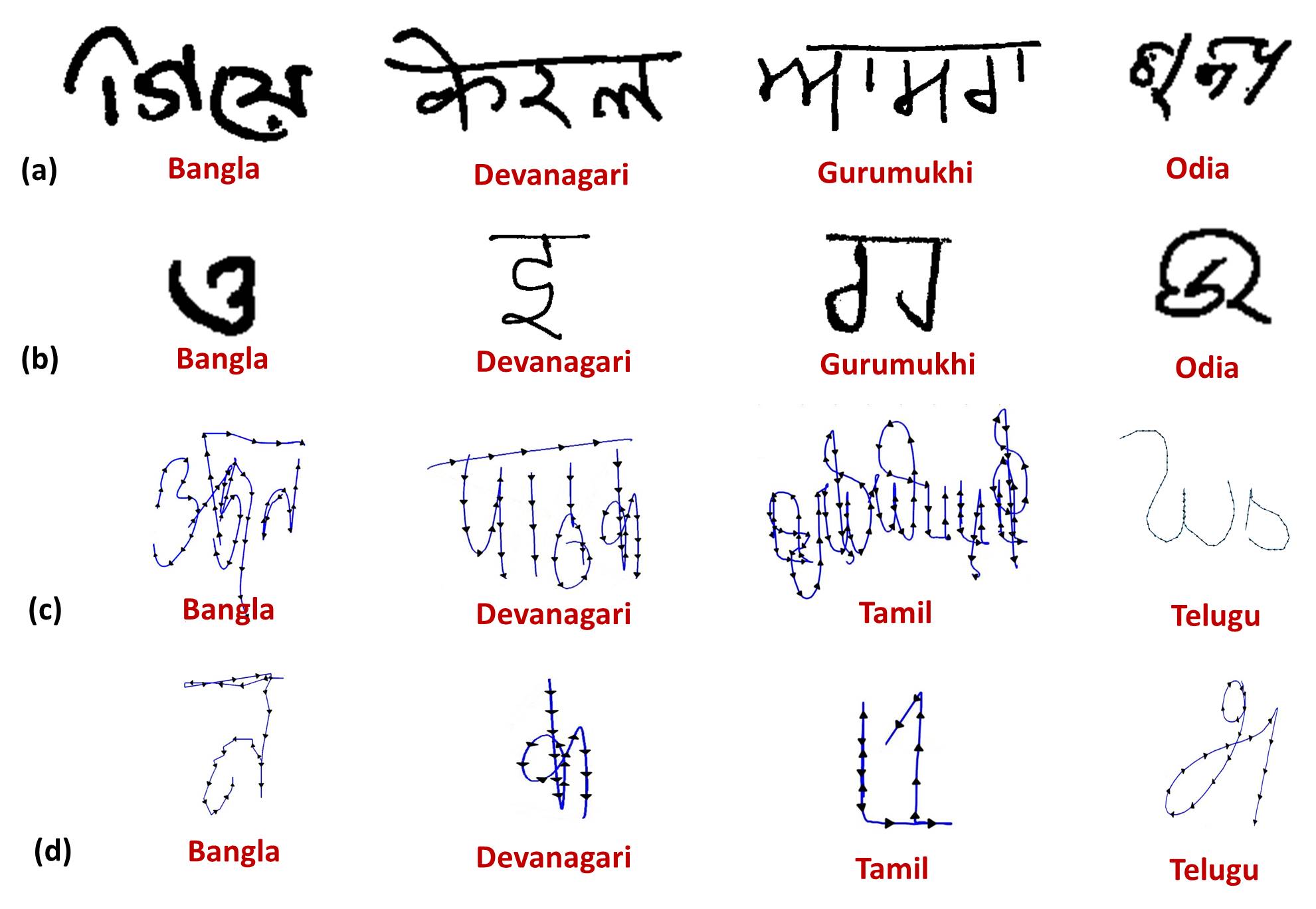}}
	\end{center}
	\centering
	\caption{Examples showing different samples from our Indic script identification dataset}
	\label{fig:fig14}
\end{figure}

\subsection{Different Baselines}
As mentioned earlier, there is no such earlier framework for handwritten script identification using multi-modal deep network which can perform for both offline and online data simultaneously using a single model. However, one of the naive approaches is to convert the data from either modality to its equivalent opposite modality, and feed the required modality to the framework based on the modality it has been trained on. For instance, if a framework has been trained using offline handwritten data, and we have online handwritten data for testing. In this case, the naive approach is to convert online data to its offline equivalent and test using the model trained from offline handwritten data. However, there is a major limitation of using such naive approach. Although, handwritten data can be converted between two modalities, the data distribution of converted data is not similar to the real one, thus limiting the performance. In order to justify the superiority of our method, we evaluated both in-modality and cross-modality performance for all the baseline methods. Another contribution of our framework is that our framework can be trained with light weight character level data and can achieve performance as that of traditional way of training a script identification model using word level data. Hence, we report the performance of our framework for both character level and word level data for training. To perform a fair comparison between word level and character level data for training, we use the nearly equal number of word and character level data for training. In Table \ref{tab1}, we have mentioned the number of data for word level and character level data from offline and online modalities are present; and it shows the number of sample is nearly with in a same range. For every experiment, cross modality(training from online data and testing on offline data or vice-versa) or cross level(i.e. training from character level data and testing on word level data or vice-versa), we use 7 fold data of a particular level or modality for training, and  testing and validation have been done on the data of other modality or level with 2 and 1 fold each, respectively.

To compare our proposed framework, we have defined a few base line methods based on deep neural network architecture. We have justified the limitations of every baseline with respect to our proposed framework. All the base line methods are defined for single modality data in order to justify the improvement in performance, we achieve due to our multi-modal framework. Also, our proposed novel multi-modal fusion method is compared with different traditional multi-modal fusion methods in section 4.3. The base line methods using different traditional classifiers and hand-crafted features are detailed in section 4.5 separately. For the first two baseline methods, we have just sliced the offline and online stream of our multi-modal network into two different baselines for online and offline data respectively. The performance comparison with these two baseline justifies the necessity of designing a multi-modal deep framework. 

\paragraph{DL\_1:}For this baseline, we use 1D-convolutional-LSTM network for online handwritten script identification. This is the same configuration as used for our online stream network and is only trained from online data. The softmax classification layer is used at the output of last time step. Although the performance is competitive in case of online data, the major limitation is, the cross-modality performance is restricted since the network is unaware of the data distribution of offline data.

\paragraph{DL\_2:}For this baseline, we use 2D-convolutional-LSTM network for offline handwritten script identification identification following the same architecture as mentioned for offline stream of our network. This has been trained only from the offline images. Here also, the main limitation is that it does not perform well for cross-modal data.

\paragraph{DL\_3:} One of the important contributions in our online stream network is the use of 1D-convolutional network over the online coordinate points in order to learn the structural correlation of among neighboring pixel points. Hence, we define our third base line in order to justify the improvement, we achieve due to use of this 1D-convolutional network. Here, we directly feed the coordinate points into the LSTM network for script identification and evaluate the identification performance. The architecture and setup for different hyper parameters are kept same as mentioned in section \ref{impl}. 

\begin{figure}[!h!t!b]
	\begin{center}
		\fbox{\includegraphics[scale = 0.50]{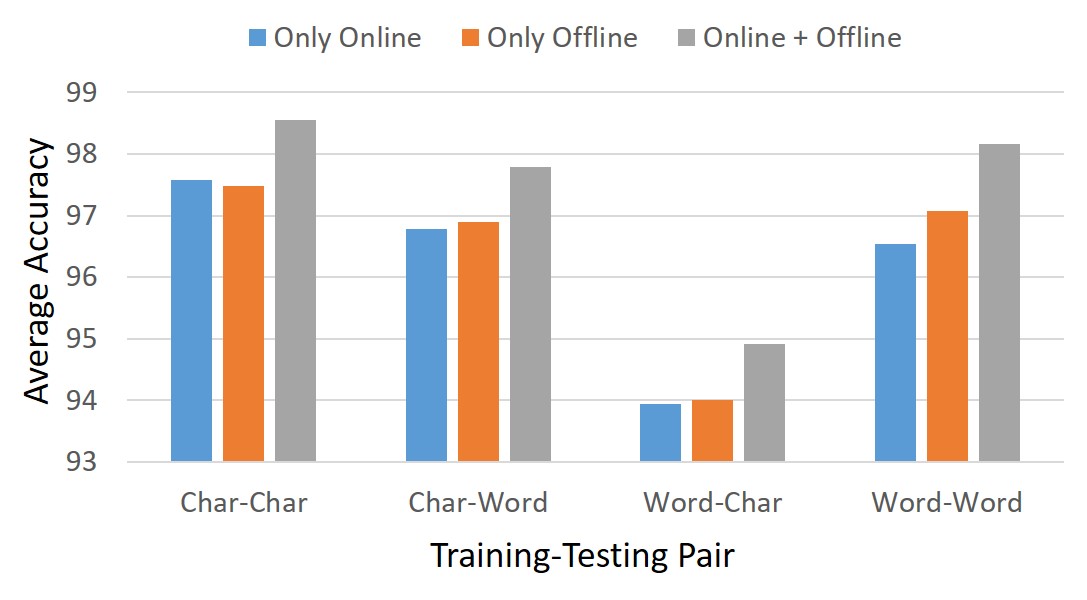}}
	\end{center}
	\centering
	\caption{Comparative study of using both offline-online modality and single modality at a time for training}
	\label{fig:fig11}
\end{figure}

\begin{figure}[!h!t!b]
	\begin{center}
		\fbox{\includegraphics[scale = 0.50]{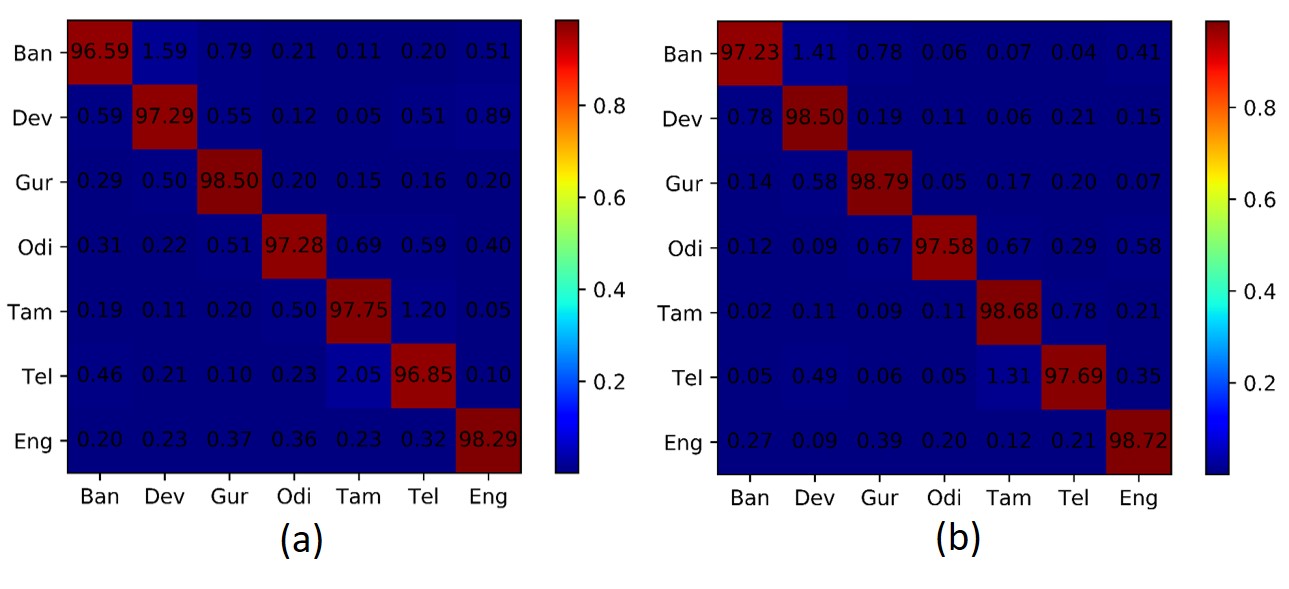}}
	\end{center}
	\centering
	\caption{Confusion matrix for (a) character level training and (b) word level training. Testing is done on word level for both the experiments. First three letters of the script name are used as the acronym.}
	\label{fig:fig12}
\end{figure}

Our framework is trained from data in paired modality, where the original modality is present from both offline and online modality with equal distribution in the training data. We have evaluated the performance our model for both real online and offline data individually. We found that our model generalizes well for both online and offline data with no significant change in the accuracy. The results are reported in Table 2. However, we have evaluated the performance using only one modality as the source of data for training. It has been observed that the performance of script identification decreases in this case compared to training the network using both the modalities as the source. Hence, we conclude that our proposed architecture generalizes well for both the modalities when it is trained from both online and offline modality. A comparative study has been shown in Fig. \ref{fig:fig11}. Interestingly, note that,  model trained from only online-data performs poorly than offline in case of last three combinations (char-word, word-char and word-word), though difference is marginal. The reasoning behind such observation could be (a) while we redraw offline version from online data, we connect the successive points, followed by a morphological thickening operation. Following Figure \ref{fig:fig8} and \ref{fig:fig14}, there exist some noticeable difference between redrawn and original offline images,  mainly in terms of non-uniform thickness of the stroke or random jittering in the shape of strokes in case of real offline images, compared to  uniform and smooth stroke width of redrawn sample. Owing to this domain gap, the trained model from redrawn offline image (for offline branch) could not generalize well for original offline image of testing set. (b) Although online modality is claimed to perform better for handwriting recognition, this may not be always true for script recognition in Indic script. Due to complex nature of indic script’s characters, presence of matra, and compound character, stroke sequence of a word  could be written in different ways, therefore it could be challenging if some unknown writing stroke sequence appears in testing data. The confusion matrix for both character and word level training is shown in Figure \ref{fig:fig12}.

% \usepackage{multirow}
\begin{table}[]
\centering
\caption{Script identification performance using different baseline schemes}
\label{my-label}
\begin{tabular}{|c|c|c|c|c|}
\hline
Method                                                                                                & \begin{tabular}[c]{@{}c@{}}Training \\ Data\end{tabular} & \begin{tabular}[c]{@{}c@{}}Testing \\ Data\end{tabular} & \begin{tabular}[c]{@{}c@{}}With in modality\\ Accuracy\end{tabular} & \begin{tabular}[c]{@{}c@{}}Cross modality\\ Accuracy\end{tabular} \\ \hline
\multirow{4}{*}{\begin{tabular}[c]{@{}c@{}}DL\_1\\ (Training from\\ Online Data)\end{tabular}}  & Character                                                      & Character                                                     & 95.61                                                               & 90.94                                                             \\ \cline{2-5} 
                                                                                                      & Character                                                      & Word                                                          & 94.84                                                               & 89.84                                                             \\ \cline{2-5} 
                                                                                                      & Word                                                           & Character                                                     & 92.13                                                               & 87.67                                                             \\ \cline{2-5} 
                                                                                                      & Word                                                           & Word                                                          & 95.14                                                               & 90.12                                                             \\ \hline
\multirow{4}{*}{\begin{tabular}[c]{@{}c@{}}DL\_2\\ (Training from\\ Offline Data)\end{tabular}} & Character                                                      & Character                                                     & 95.47                                                               & 90.96                                                             \\ \cline{2-5} 
                                                                                                      & Character                                                      & Word                                                          & 94.59                                                               & 90.04                                                             \\ \cline{2-5} 
                                                                                                      & Word                                                           & Character                                                     & 91.83                                                               & 87.23                                                             \\ \cline{2-5} 
                                                                                                      & Word                                                           & Word                                                          & 95.31                                                               & 90.54                                                             \\ \hline
\multirow{4}{*}{\begin{tabular}[c]{@{}c@{}}DL\_3\\ (Training from \\ Online Data)\end{tabular}} & Character                                                      & Character                                                     & 92.48                                                               & 87.47                                                             \\ \cline{2-5} 
                                                                                                      & Character                                                      & Word                                                          & 91.43                                                               & 86.64                                                             \\ \cline{2-5} 
                                                                                                      & Word                                                           & Character                                                     & 89.32                                                               & 85.14                                                             \\ \cline{2-5} 
                                                                                                      & Word                                                           & Word                                                          & 92.12                                                               & 87.34                                                             \\ \hline
\end{tabular}
\end{table}

\begin{table}[]
\centering
\caption{Script identification performance using our proposed framework}
\label{tab:Acc}
\begin{tabular}{|c|c|c|c|c|}
\hline
\begin{tabular}[c]{@{}c@{}}Training\\ Data Level\end{tabular} & \begin{tabular}[c]{@{}c@{}}Testing\\ Data Level\end{tabular} & \begin{tabular}[c]{@{}c@{}}Online Modality\\ Accuracy\end{tabular} & \begin{tabular}[c]{@{}c@{}}Offline Modality\\ Accuracy\end{tabular} & \begin{tabular}[c]{@{}c@{}}Average \\ Accuracy\end{tabular} \\ \hline
Character                                                     & Character                                                    & 98.59                                                              & 98.48                                                               & 98.55                                                       \\ \hline
Character                                                     & Word                                                         & 97.75                                                              & 97.81 & 97.79                                                       \\ \hline
Word                                                          & Character                                                    & 94.89                                                              & 94.91                                                               & 94.92                                                       \\ \hline
Word                                                          & Word                                                         & 98.11                                                              & 98.23                                                               & 98.17                                                       \\ \hline
\end{tabular}
\end{table}

\subsection{Comparative study with different multi-modal fusion methods}

One of most popular application of multi-modal fusion approaches is Visual Question Answering\citep{ilievski2017multimodal}, where image and language feature  representation are combined. Here, we also combine the offline and online feature representations for script identification using a conditional multi-modal fusion method. We compare our multi-modal fusion method with different traditional fusion methods popular in the literature. Lets denote the feature representation of offline and online modality as $F_{1}$ and $F_{2}$, both of which is of dimension $\mathbb{R}^{1\times K}$.The traditional approaches used in our comparison are as follows: Firstly, One of the approaches is to concatenate these two features $F_{1}$ and $F_{2}$ and feed it to a fully connected and a softmax layer for classification. Secondly, $F_{1}$ and $F_{2}$ can be added or multiplied element wise followed by a fully connected and a softmax layer for classification. Thirdly, we have considered the outer product between $F_{1}$ and $F_{2}$ and bilinear pooling \citep{lin2015bilinear} followed by a fully connected and a softmax layer for final classification. Fourthly, we use multi-modal Compact Bilinear Pooling with pooling dimension 4K for multi-modal fusion. More details about multi-modal Compact Bilinear(MCB) Pooling can be found in \citep{gao2016compact}. We have also evaluated the performance of our multi-modal fusion method with out using conditional fusion. The results for different multi-modal fusion strategies are reported in Table \ref{fusion}.

\begin{table}[]
\centering
\caption{Comparative study with different multi-modal fusion strategy}
\label{fusion}
\begin{tabular}{|c|c|c|c|c|}
\hline
                                                                                      & \multicolumn{4}{c|}{Training-Testing Pair Combination} \\ \hline
Method                                                                                & Char-Char    & Char-Word   & Word-Char   & Word-Word   \\ \hline
Elementwise Sum                                                                       & 96.12        & 95.61       & 92.87       & 95.94       \\ \hline
Concatenation                                                                         & 96.81        & 96.34       & 93.37       & 96.64       \\ \hline
Elementwise Product                                                                   & 96.17        & 95.57       & 93.01       & 96.05       \\ \hline
\begin{tabular}[c]{@{}c@{}}Outer Product + Bilinear\\   Pooling\end{tabular}          & 96.94        & 96.59       & 93.61       & 96.83       \\ \hline
MCB Pooling                                                                           & 97.14        & 96.72       & 93.87       & 97.03       \\ \hline
\begin{tabular}[c]{@{}c@{}}Proposed Fusion Method \\ (Without Condition)\end{tabular} & 97.89        & 97.08       & 94.27       & 97.49       \\ \hline
\begin{tabular}[c]{@{}c@{}}Proposed Fusion Method\\ (With Condition)\end{tabular}     & 98.55        & 97.79    & 94.92       & 98.17       \\ \hline
\end{tabular}
\end{table}

%\subsection{Analysis of Character Level Training}

%In this paper, we hold the opinion that character level training data is sufficient enough to achieve state-of-the-art script identification performance as that of training from word level data. In addition, character level training posses a few advantages as mentioned in section 1. In Figure \ref{fig:fig13}, a comparative study has been performed between character and word level training with respect to varying number of training sample. From the results, it is justified that comparatively fewer character level training data is sufficient to achieve the optimum performance compared to word level training data. Word level training outperforms character level training method after a certain number of training samples. However, the difference between character and word level training is almost negligible with some added advantages on the side of character level training.  

%\begin{figure}[!h!t!b]
%	\begin{center}
%		\fbox{\includegraphics[scale = 0.650]{images/fig13.jpg}}
%	\end{center}
%	\centering
%	\caption{Comparative study between character and word level training with respect to varying number of training sample. Accuracy is reported for word level testing.}
%	\label{fig:fig13}
%\end{figure}

\subsection{Comparative study with different traditional classifiers}

In this paper, we have considered convolutional layers stacked with LSTMs for due to their superior performance reported in recent literature. It is a deep multi-modal network which can be trained in a end to end manner using back propagation and it takes input for both offline and online modality of the data. In recent literature, Deep Neural Network has achieved a great superiority over the traditional classifiers and handcrafted features.  Most of the previous methods for handwritten script identification are based on this traditional classifiers and handcrafted features.  However, since there is no standard dataset, we can not compare our deep learning based system with those methods. Hence, we employed some baseline methods using some popular traditional classifiers and popular handcrafted features. We name these methods as Traditional Approaches(TA) and report the results in our 7 handwritten scripts dataset. We show the performance using both character level and word level data as training. Also, we report the cross modality script identification accuracy. However, training has been done from single modality, i.e. no multi-modal combination, for these traditional methods.  The different baseline approaches considered in our experiments are described in the following subsections.

\subsubsection{Performance on Offline Data}\

For offline word images, we have used PHOG (Pyramidal Histogram of Oriented Gradient) and LBP (Local Binary Pattern) for feature extraction. PHOG is gradient based feature which has been utilized in sliding window based handwriting recognition task \cite{roy2016hmm, BHUNIA201812}. LBP is a texture descriptor which has been further used in different tasks like, facial expression recognition \citep{moore2011local}, handwritten word spotting in historical documents \citep{dey2016local} etc.

\paragraph{TA\_1:} For this baseline, we have used two popular state-of the art traditional classifiers \citep{fernandez2014we}, SVM(Support Vector Machine) and Random Forest, for the script identification task along with PHOG and LBP features, respectively. Both SVM and Random Forest have been studied in various classification problems extensively in both Computer Vision and Document Image Analysis communities. Here, we have extracted the handcrafted feature using PHOG and LBP respectively, and evaluate the classification performance using SVM and Random Forest classifiers, respectively. The results are reported in Table \ref{tab5} as TA\_1. It is to be noted that using this baseline, it is not possible to predict the script at word level using character level training data, since SVMs or Random Forests are not good in handling sequential data. Hence, the results are reported accordingly. %%%%

\paragraph{TA\_2:} In this baseline, we have explored the sequence-based script identification approach using two traditional sequential classifiers HMM and HCRF, respectively. For sequential feature extraction from word or character level images, a sliding window moves from left to right of the image and PHOG or LBP feature is extracted from each sliding window, which denotes the feature at single time step. This type of sliding window based classification approach has been used in music score writer identification \citep{roy2017hmm}. As we are dealing with sequence based classification approach in this baseline, it is possible to use character level data for training in order to predict the script at word level. The results are reported in Table \ref{tab5} as TA\_2.

\paragraph{TA\_3:} Here, we use the same sliding window based feature extraction approach as used in TA\_2, the only difference is that we use 2 layers LSTM as the classifier. Performance of different handcrafted features with LSTM has been studied in \citep{chherawala2016feature} for offline handwriting recognition. Hence, we employ a similar framework for the script identification task, and name it as TA\_3. Results of each corresponding experiment has been reported in Table \ref{tab6}

% \usepackage{multirow}
\begin{table}[]
\centering
\caption{Comparative study with different traditional baseline methods}
\label{tab5}
\begin{tabular}{|c|c|c|c|c|c|c|}
\hline
Method                  & Classifier                                                                & Feature               & \begin{tabular}[c]{@{}c@{}}Training \\ Data Level\end{tabular} & \begin{tabular}[c]{@{}c@{}}Testing \\ Data Level\end{tabular} & \begin{tabular}[c]{@{}c@{}}With in\\ Modality Acc.\end{tabular} & \begin{tabular}[c]{@{}c@{}}Cross \\ Modality Acc.\end{tabular} \\ \hline
\multirow{12}{*}{TA\_1} & \multirow{6}{*}{SVM}                                                      & \multirow{3}{*}{LBP}  & Character                                                          & Character                                                     & 90.78                                                           & 85.48                                                          \\ \cline{4-7} 
                        &                                                                           &                       & Word                                                           & Character                                                     & 87.29                                                           & 83.12                                                          \\ \cline{4-7} 
                        &                                                                           &                       & Word                                                           & Word                                                          & 89.44                                                           & 84.69                                                          \\ \cline{3-7} 
                        &                                                                           & \multirow{3}{*}{PHOG} & Character                                                      & Character                                                     & 91.48                                                           & 85.78                                                          \\ \cline{4-7} 
                        &                                                                           &                       & Word                                                           & Character                                                     & 88.32                                                           & 83.67                                                          \\ \cline{4-7} 
                        &                                                                           &                       & Word                                                           & Word                                                          & 90.49                                                           & 85.14                                                          \\ \cline{2-7} 
                        & \multirow{6}{*}{\begin{tabular}[c]{@{}c@{}}Random\\  Forest\end{tabular}} & \multirow{3}{*}{LBP}  & Character                                                      & Character                                                     & 90.87                                                           & 85.94                                                          \\ \cline{4-7} 
                        &                                                                           &                       & Word                                                           & Character                                                     & 87.21                                                           & 83.64                                                          \\ \cline{4-7} 
                        &                                                                           &                       & Word                                                           & Word                                                          & 89.51                                                           & 84.47                                                          \\ \cline{3-7} 
                        &                                                                           & \multirow{3}{*}{PHOG} & Character                                                      & Character                                                     & 91.54                                                           & 86.71                                                          \\ \cline{4-7} 
                        &                                                                           &                       & Word                                                           & Character                                                     & 88.57                                                           & 84.39                                                          \\ \cline{4-7} 
                        &                                                                           &                       & Word                                                           & Word                                                          & 90.42                                                           & 85.49                                                          \\ \hline
\multirow{16}{*}{TA\_2} & \multirow{8}{*}{HMM}                                                      & \multirow{4}{*}{LBP}  & Character                                                      & Character                                                     & 91.48                                                           & 86.65                                                          \\ \cline{4-7} 
                        &                                                                           &                       & Character                                                      & Word                                                          & 90.53                                                           & 85.17                                                          \\ \cline{4-7} 
                        &                                                                           &                       & Word                                                           & Character                                                     & 88.51                                                           & 84.74                                                          \\ \cline{4-7} 
                        &                                                                           &                       & Word                                                           & Word                                                          & 91.11                                                           & 86.39                                                          \\ \cline{3-7} 
                        &                                                                           & \multirow{4}{*}{PHOG} & Character                                                      & Character                                                     & 91.94                                                           & 86.46                                                          \\ \cline{4-7} 
                        &                                                                           &                       & Character                                                      & Word                                                          & 90.91                                                           & 85.37                                                          \\ \cline{4-7} 
                        &                                                                           &                       & Word                                                           & Character                                                     & 89.14                                                           & 84.69                                                          \\ \cline{4-7} 
                        &                                                                           &                       & Word                                                           & Word                                                          & 91.81                                                           & 86.44                                                          \\ \cline{2-7} 
                        & \multirow{8}{*}{HCRF}                                                     & \multirow{4}{*}{LBP}  & Character                                                      & Character                                                     & 91.34                                                           & 86.46                                                          \\ \cline{4-7} 
                        &                                                                           &                       & Character                                                      & Word                                                          & 90.37                                                           & 85.71                                                          \\ \cline{4-7} 
                        &                                                                           &                       & Word                                                           & Character                                                     & 88.97                                                           & 84.72                                                          \\ \cline{4-7} 
                        &                                                                           &                       & Word                                                           & Word                                                          & 91.23                                                           & 86.82                                                          \\ \cline{3-7} 
                        &                                                                           & \multirow{4}{*}{PHOG} & Character                                                      & Character                                                     & 91.87                                                           & 86.55                                                          \\ \cline{4-7} 
                        &                                                                           &                       & Character                                                      & Word                                                          & 90.84                                                           & 85.67                                                          \\ \cline{4-7} 
                        &                                                                           &                       & Word                                                           & Character                                                     & 89.34                                                           & 85.41                                                          \\ \cline{4-7} 
                        &                                                                           &                       & Word                                                           & Word                                                          & 91.97                                                           & 86.91                                                          \\ \hline
\end{tabular}
\end{table}

% \usepackage{multirow}
\begin{table}[]
\centering
\caption{Comparative study with different traditional baseline methods}
\label{tab6}
\begin{tabular}{|c|c|c|c|c|c|c|}
\hline
Method                   & Classifier            & Feature                 & \begin{tabular}[c]{@{}c@{}}Training \\ Data Level\end{tabular} & \begin{tabular}[c]{@{}c@{}}Testing \\ Data Level\end{tabular} & \begin{tabular}[c]{@{}c@{}}With-in \\ Modality Acc.\end{tabular} & \begin{tabular}[c]{@{}c@{}}Cross \\ Modality Acc\end{tabular} \\ \hline
\multirow{8}{*}{TA\_3} & \multirow{8}{*}{LSTM} & \multirow{4}{*}{LBP}    & Character                                                      & Character                                                     & 91.64                                                            & 86.47                                                         \\ \cline{4-7} 
                         &                       &                         & Character                                                      & Word                                                          & 90.54                                                            & 85.69                                                         \\ \cline{4-7} 
                         &                       &                         & Word                                                           & Character                                                     & 88.87                                                            & 84.03                                                         \\ \cline{4-7} 
                         &                       &                         & Word                                                           & Word                                                          & 91.63                                                            & 86.23                                                         \\ \cline{3-7} 
                         &                       & \multirow{4}{*}{PHOG}   & Character                                                      & Character                                                     & 92.11                                                            & 87.39                                                         \\ \cline{4-7} 
                         &                       &                         & Character                                                      & Word                                                          & 91.01                                                            & 86.46                                                         \\ \cline{4-7} 
                         &                       &                         & Word                                                           & Character                                                     & 89.31                                                            & 84.97                                                         \\ \cline{4-7} 
                         &                       &                         & Word                                                           & Word                                                          & 91.82                                                            & 86.89                                                         \\ \hline
\multirow{8}{*}{TA\_4} & \multirow{4}{*}{HMM}  & \multirow{4}{*}{NPEN++} & Character                                                      & Character                                                     & 92.34                                                            & 84.44                                                         \\ \cline{4-7} 
                         &                       &                         & Character                                                      & Word                                                          & 91.47                                                            & 83.69                                                         \\ \cline{4-7} 
                         &                       &                         & Word                                                           & Character                                                     & 88.94                                                            & 81.41                                                         \\ \cline{4-7} 
                         &                       &                         & Word                                                           & Word                                                          & 92.03                                                            & 83.87                                                         \\ \cline{2-7} 
                         & \multirow{4}{*}{HCRF} & \multirow{4}{*}{NPEN++} & Character                                                      & Character                                                     & 92.31                                                            & 84.47                                                         \\ \cline{4-7} 
                         &                       &                         & Character                                                      & Word                                                          & 91.37                                                            & 83.41                                                         \\ \cline{4-7} 
                         &                       &                         & Word                                                           & Character                                                     & 88.67                                                            & 81.67                                                         \\ \cline{4-7} 
                         &                       &                         & Word                                                           & Word                                                          & 92.08                                                            & 83.69                                                         \\ \hline
\multirow{4}{*}{TA\_5} & \multirow{4}{*}{LSTM} & \multirow{4}{*}{NPEN++} & Character                                                      & Character                                                     & 92.54                                                            & 84.89                                                         \\ \cline{4-7} 
                         &                       &                         & Character                                                      & Word                                                          & 91.67                                                            & 83.85                                                         \\ \cline{4-7} 
                         &                       &                         & Word                                                           & Character                                                     & 89.05                                                            & 81.09                                                         \\ \cline{4-7} 
                         &                       &                         & Word                                                           & Word                                                          & 92.17                                                            & 84.12                                                         \\ \hline
\end{tabular}
\end{table}

\subsubsection{Performance on Online Data}

Online data contains the successive coordinate points representing the trajectory of pen's movement. For online handwriting recognition, one of the most popular handcrafted feature descriptor is NPEN++ \citep{jaeger2001online}. We have considered this feature in our traditional baseline method for handwritten online script identification. Here, five different features namely, Curliness (CR), Writing direction (WD), Linearity (LR), Slope (SP), and  Curvature (CV) are extracted from the sequence of coordinate points and are concatenated to form the final feature descriptor. More details about these features can be found in \citep{jaeger2001online}.

\paragraph{TA\_4:} In this baseline, we have used the NPEN++ feature along with sequential classifier HMM and HCRF for each case. HMM and NPEN++ feature have been used in online handwriting recognition in \citep{jaeger2001online}. Hence, it is reasonable to evaluate the performance of these combinations for online handwritten script identification task. The results are reported in Table \ref{tab6} as TA\_4. 

\paragraph{TA\_5:} In this framework, we directly feed the online sequential coordinate point in a LSTM network for script identification. This type of approach has been addressed in the literature for handwriting recognition in \citep{graves2008unconstrained}. We report the result for this approach in Table \ref{tab6}.

\subsection{Performance on Public PHDIndic Dataset}
\begin{figure}[!h!t!b]
	\begin{center}
		\fbox{\includegraphics[scale = 0.6]{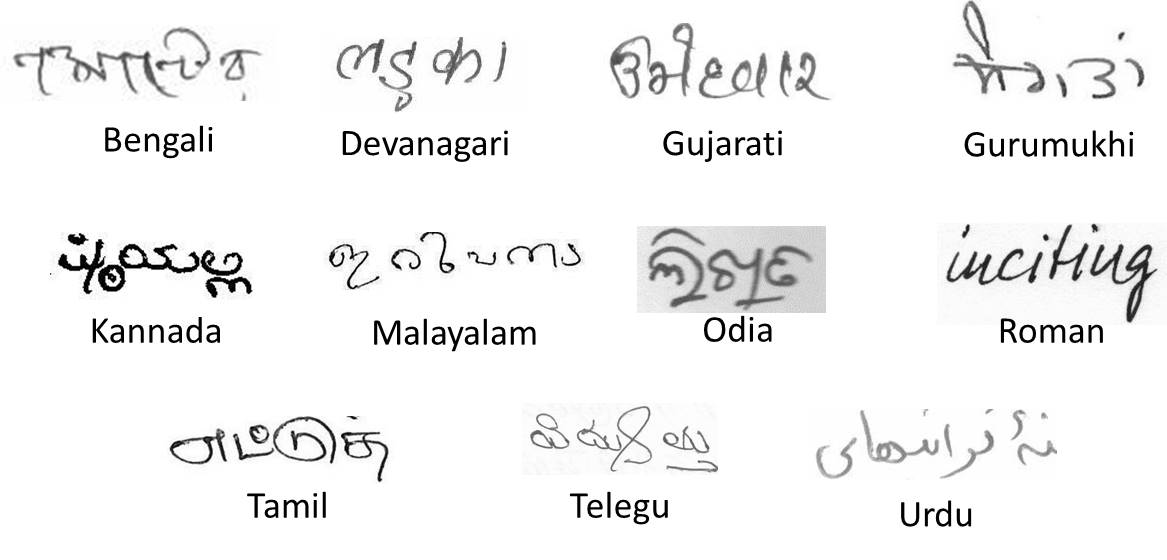}}
	\end{center}
	\centering
	\caption{Sample word-level images from different scripts of PHDIndic~\cite{obaidullah2018phdindic_11} dataset.}
	\label{fig:phdindic}
\end{figure}

Recently, authors in \cite{obaidullah2018phdindic_11} introduced a new dataset for handwritten Indic script identification, which consists of 1000 words each from 11 different scripts, namely Bengali, Devanagari, Gujarati, Gurumukhi, Kannada, Malayalam, Odia, Roman, Tamil, Telegu, Urdu. Some word level samples are shown in Figure \ref{fig:phdindic}. Note that PHDIndic is completely a word-level offline dataset and no online dataset is available for all the 11 scripts. Please also note that our framework is designed with a motivation to work both for offline and online data parallelly. In order to justify the contribution of our multi-modal framework, we need such a dataset which has both offline and online data for all the scripts, so that we can evaluate within modality and cross modality performance that would explain the contribution of different design choices in our network architecture. Nevertheless, we have evaluated the performance on offline word-level data from PHDIndic dataset in order to validate our framework on a publicly available dataset. Following~\cite{ukil2018deep}, we use 80\% images from each script as training and rest 20\% for testing. For the online stream, we feed online-converted data from offline images. We achieved 94.74\% accuracy on the testing set. As shown in Figure \ref{fig:fig11}, that our framework generalizes well while we train it from both real offline-online data. Therefore, had we been able to train the model also including real online data for all the 11 scripts, the accuracy is supposed to get improved. Moreover, our method is competitive with other recently introduced frameworks \cite{obaidullah2018phdindic_11, ukil2019improved}. The foremost thing here is to noted that all these frameworks  \cite{obaidullah2018phdindic_11, ukil2019improved} have been designed explicitly for offline data which has no way to handle online data equally; however our framework can handle both offline and online data parallely with nearly equal efficiency (see Table \ref{tab:Acc}). 

\subsection{Error Analysis}

\begin{figure}[!h!t!b]
	\begin{center}
		\fbox{\includegraphics[scale = 0.5]{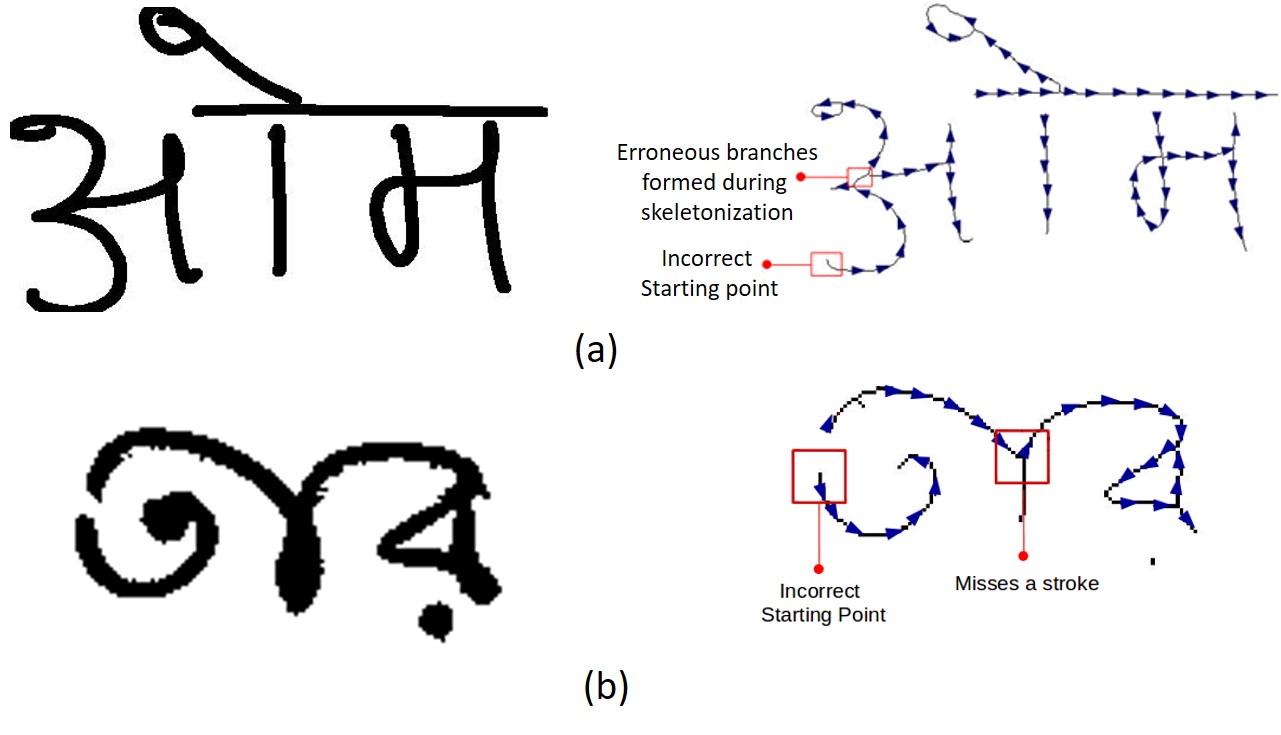}}
	\end{center}
	\centering
	\caption{ Errors in stroke recovery method}
	\label{fig:fig10}
\end{figure}

Inter modality conversion is one of the main crucial steps in our framework, since our proposed architecture takes offline-online modality pair of a data sample as the input. Although the conversion of online  to offline modality is a trivial and simple one, the recovery of stroke information form offline data is a bit challenging due to free flow nature of handwriting by different individuals. For this task, we have adopted the method in \citep{nel2005estimating}. However, during our experimental analysis, we have observed that the offline to online conversion algorithm fails to recover proper stroke sequences in some specific cases. Figure \ref{fig:fig10} shows some examples, where the the our adopted algorithm fails to recover the expected stroke sequence in some specific cases. One of the main reasons of this problem is skeletonization error that appears as unwanted skeleton branches with incorrect angles due to uneven thickness of the handwritten data and surface noise. It is also observed from 
the second image of Figure \ref{fig:fig10} that the algorithm tends to miss strokes at the junction points due to the presence of the Matra which is very common feature among Indic scripts. This problem can be solved to some extent by Matra removal as proposed by \citep{roy2016hmm}.

\section{Conclusions and Future Work}
In this paper we have proposed a new method for script identification which has provided us with satisfactory results. Handwritten text present in either modality, online or offline has been received and the absent modality has been recreated using inter modality conversion. After such recreation, both modalities had been fed in pair into a deep neural network. This designed neural network uses both sets of information from both modalities to employ multi-modal fusion thus combining the features adaptively. Two significant achievements obtained are the designing of one single training model which encompasses the training of both modalities of data. Secondly, the features of two modalities are combined to produce more accurate results. 

As evident from the results, our method performs better than almost every other state-of-the art method for handwritten script identification. A few drawbacks include incomplete conversion of  modality to the other, but the conditions existent for such cases are rare. The offline to online conversion can be handled using a deep end-to-end network \cite{kumarbhunia2018handwriting} and can be included as a sub-module in our architecture. Our future work would include fine tuning our proposed deep network architecture and including more Indic scripts to increase the scope of application of our method. One of the most promising future research directions would be to design a single deep model for offline and online handwriting recognition in a single deep neural network through exploring information from both the modalities. 

\bibliographystyle{elsarticle-num}
\bibliography{mybibfile}

\begin{thebibliography}{10}
\bibitem{shi2016end}
B.~Shi, X.~Bai, C.~Yao, An end-to-end trainable neural network for image-based
  sequence recognition and its application to scene text recognition, IEEE
  transactions on pattern analysis and machine intelligence.
\end{thebibliography}

\end{document}